\documentclass[11pt]{article}

\usepackage[T1]{fontenc}
\usepackage[utf8]{inputenc}
\usepackage[a4paper,margin=1in]{geometry}
\usepackage{microtype} 

\usepackage{amsmath,amssymb,amsfonts}
\numberwithin{equation}{section}

\usepackage{graphicx}
\graphicspath{{figures/}}
\usepackage{booktabs}
\usepackage{caption}
\usepackage{subcaption}
\usepackage{float}

\usepackage[numbers,sort&compress]{natbib}
\usepackage{url}
\usepackage{hyperref}
\usepackage[capitalise,noabbrev]{cleveref}

\usepackage{authblk}
\usepackage{orcidlink}

\title{Component-Adaptive and Lesion-Level Supervision for Improved Small Structure Segmentation in Brain MRI}
\author[1]{Minh Sao Khue Luu\,\orcidlink{0000-0002-6766-4966}}
\author[1]{Evgeniy N. Pavlovskiy\,\orcidlink{0000-0001-6976-1885}}
\author[1]{Bair N. Tuchinov\,\orcidlink{0000-0002-8931-9848}}
\affil[1]{The Artificial Intelligence Research Center of Novosibirsk State University, Novosibirsk 630090, Russia}
\affil[ ]{\texttt{luumsk@gmail.com}}
\date{}

\begin{document}
\maketitle

\begin{abstract}
Small lesions in brain magnetic resonance imaging (MRI) are easily missed: a segmentation model can reach a high Dice similarity coefficient (DSC) while leaving many of them undetected. We ask whether this omission comes from the network or from the training objective. In multiple sclerosis lesion segmentation, nnU-Net trained with its default Dice and cross-entropy loss produces almost no response to most of the small lesions it misses, so no decision threshold recovers them, and four other architectures trained with the same loss detect no more of them. We therefore change only the objective. CATMIL adds two lesion-level terms to the nnU-Net loss: a Component-Adaptive Tversky (CAT) term, which gives each lesion a weight nearly independent of its size, and a Multiple Instance Learning (MIL) term, which treats each lesion as a bag of voxels and penalizes any lesion in which no voxel is detected. On MSLesSeg, CATMIL raises small-lesion recall from 0.796 to 0.873 (difference 95\% CI +0.030 to +0.157) and misses 48\% fewer lesions, at comparable DSC. The gain holds on the cross-validation predictions of 47 further patients, for lesions of at least 3~mm, and on a second dataset trained with the same loss weights. CATMIL also detects more small lesions than CC-DiceCE, an instance-level loss that balances lesions within an overlap term (0.873 vs.\ 0.831), indicating that lesions must be supervised not only in proportion but also for their detection. The cost is additional small false-positive components; a component-size filter removes most of them, and at the lesion-wise precision of the default loss CATMIL still detects more small lesions. Code: \url{https://github.com/luumsk/SmallLesionMRI}.

\end{abstract}

\paragraph{Keywords:} brain MRI; component-adaptive loss; multiple instance learning; small lesions segmentation

\section{Introduction}
\label{sec:intro}
Small lesions in brain magnetic resonance imaging (MRI) are sparse targets
embedded in a large volume of normal-appearing tissue. In multiple
sclerosis (MS), the number of lesions and the appearance of new lesions on
follow-up scans are central to diagnosis~\cite{barkhof_2024_2025} and to
the monitoring of disease activity~\cite{wattjes_2021_2021}, so a small
lesion that is never detected is a clinically relevant error, even when it
barely changes the segmented volume.

Progress in medical image segmentation has come largely from architectures
and pipelines: U-Net and its 3D extensions~\cite{navab_u-net_2015}, the
self-configuring nnU-Net~\cite{isensee_nnu-net_2021}, and transformer-based
and hybrid models~\cite{hatamizadeh_unetr_2021,hatamizadeh_swin_2022}. Yet
small lesions are still missed by state-of-the-art models; in the
nnU-Net-based FLAMeS, most missed lesions were smaller than
10~mm$^3$~\cite{dereskewicz_flames_2025}. This raises a question that
architectural progress does not answer: are small lesions missed because
the network cannot represent them, or because the training objective gives
it no reason to?

The objective is a plausible cause. In a voxel-wise loss, a lesion of a
few voxels is a negligible part of the objective, so a model can reach a
high Dice similarity coefficient (DSC) while ignoring it. Losses for class
imbalance reweight voxels, and instance-level losses balance lesions of
different sizes, but both still judge each lesion by how much of it is
overlapped. Neither asks the more basic question of whether the lesion is
detected at all, and no existing objective supervises this separately from
delineation (\cref{sec:related}).

We test this explanation by changing only the objective. CATMIL adds two
lesion-level terms to the standard nnU-Net loss, one for each property
that a voxel-wise loss lacks: a \emph{Component-Adaptive Tversky (CAT)}
term, which gives small and large lesions comparable weight, and a
\emph{Multiple Instance Learning (MIL)} term, which penalizes any lesion in
which no voxel is detected. Because the network is fixed, differences in
behavior can be attributed to the supervision. Our contributions are:
\begin{itemize}
\item \textbf{A diagnosis of how small lesions are missed.} Standard losses
do not miss small lesions by a narrow margin: the network produces almost
no response to most of them, so the omission cannot be corrected by the
decision threshold. On MSLesSeg, four other architectures trained with the
same loss detect no more small lesions, whereas changing only the
objective recovers many of them.
\item \textbf{A lesion-level detection objective for fully supervised
segmentation.} To our knowledge, CATMIL is the first use of multiple
instance learning as a per-lesion existence constraint when voxel-level
annotations are available. We show analytically that the CAT term gives
each lesion a total weight nearly independent of its size and that the MIL
term gives every lesion the same detection signal, largest when the lesion
is missed. The objective requires no change to the network.
\item \textbf{Evidence that lesion-level supervision recovers the missed
lesions.} CATMIL detects more small lesions than losses that reweight
voxels or balance lesions within an overlap term, including the
instance-level loss CC-DiceCE, and an ablation separates the roles of its
two terms. The gain holds across patients, on 47
cross-validation patients, for lesions above the size used in clinical
reading, and on a second dataset trained with the same loss weights.
\item \textbf{A characterization of the cost.} The added sensitivity
produces small false-positive components. They can be removed after
training, and at the lesion-wise precision of the standard loss CATMIL
still detects more small lesions; against Tversky-based losses, part of
the gain reflects a more sensitive operating point.
\end{itemize}

\section{Related Work}
\label{sec:related}
The training objective for small-lesion segmentation has to satisfy two
requirements (\cref{sec:problem}): a lesion's contribution should not
scale with its size (balance), and a missed lesion should be penalized at
the lesion level (detection). This section reviews prior work according
to the extent to which it meets these requirements.

\paragraph{Voxel- and class-level reweighting.}
Early approaches primarily address the imbalance between foreground and
background. Generalised Dice loss weights each class by the inverse square
of its volume~\cite{sudre_generalised_2017}, whereas Tversky loss assigns a
greater penalty to false negatives than to false
positives~\cite{salehi_tversky_2017}. Focal loss down-weights easy
voxels~\cite{lin_focal_2017}, Focal Tversky loss combines the two
approaches~\cite{abraham_novel_2019}, and DSC++ reduces the overconfidence
associated with Dice loss~\cite{yeung_calibrating_2023}. Region-wise losses
also vary the penalty spatially, either across image
subregions~\cite{chen_adaptive_2023} or through rectified region-wise
maps~\cite{valverde_region-wise_2023}. Large-scale comparisons have found
compound losses to be the most robust choice~\cite{ma_loss_2021,azad_loss_2023}.
However, these methods weight voxels according to their class, difficulty,
or position rather than the lesion to which they belong. Consequently, a
lesion's influence on training remains proportional to its size
(\cref{eq:sl_voxel_loss}), and neither requirement is satisfied.

\paragraph{Instance-level balancing.}
A second group of methods explicitly represents lesions as connected
components. Inverse weighting~\cite{shirokikh_universal_2020} assigns each
voxel a weight inversely proportional to the volume of its lesion. When
applied to Dice, Focal, and asymmetric similarity losses, this approach
improved the detection of liver tumors, lung nodules, and brain metastases,
although delineation quality decreased on the liver dataset. Blob
loss~\cite{kofler_blob_2023}, by contrast, evaluates a base loss separately
for each lesion after masking all other lesions from the domain, and then
averages the resulting losses. With Dice as the base loss, it improved the
lesion-wise F1 score for MS lesions by 5\% relative to soft Dice loss.
CC-DiceCE~\cite{bouteille_learning_2026} builds the same kind of
per-lesion loss on the component-wise evaluation protocol of
CC-Metrics~\cite{jaus_every_2025} and achieves higher recall than blob loss,
albeit with a dataset-dependent reduction in precision. Its successor,
BiCC~\cite{bouteille_bicc_2026}, derives components from both the
annotation and the prediction, so that false-positive components also
receive an instance-level penalty, and improves lesion-wise F1 over
CC-DiceCE and blob loss on most of the evaluated datasets. Regularizing both
predictions and labels has also been proposed as a means of addressing
instance imbalance~\cite{aqib_javed_novel_2025}. These methods satisfy the
balance requirement, and the component-adaptive term introduced in this
work follows the same inverse-weighting principle. Nevertheless, they
remain overlap-based: each lesion is assessed according to the quality of
its delineation, coupling detection and delineation within a single term.
None of these methods considers only whether a lesion has been detected.
The same component-level view has been adopted in the MS domain: a blob
variant of the binary cross-entropy and Dice loss modestly improved
lesion-wise F1 and precision for cortical MS lesions with
nnU-Net~\cite{molchanova_benchmarking_2025}.

\paragraph{Lesion-level detection and multiple instance learning.}
Few loss functions directly supervise detection. ALL-Net~\cite{zhang_all-net_2021}
introduces a lesion-wise loss that models MS lesions as equal-sized spheres,
whereas ICI loss~\cite{rachmadi_improving_2024} adds a center-of-instance
term that represents each instance as a fixed-size cube centered at its
center of mass. Both methods remove lesion size from the detection signal,
but rely on surrogate shapes that the prediction must still match.
MIL~\cite{dietterich_solving_1997,ilse_attention-based_2018} expresses the
detection requirement more directly: a bag is considered positive if at
least one of its instances is positive, just as a lesion can be considered
detected if at least one of its voxels is predicted as foreground with high
confidence. In segmentation, however, MIL has been used almost exclusively
to compensate for missing voxel-level labels: pixel scores are pooled to
match image-level labels~\cite{pinheiro_image-level_2015} or bounding-box
constraints~\cite{hsu_weakly_2019}, while recent medical applications have
similarly focused on weakly supervised
segmentation~\cite{liu_object_2026,huang_geometric_2026,pal_deep_2021},
including lesion segmentation learned from image-level labels
alone~\cite{cheng_fmamil_2025}. To
the best of our knowledge, MIL has not been used as a per-lesion existence
constraint when full voxel-level annotations are available.

\paragraph{MS lesion segmentation.}
Recent advances in MS lesion segmentation have been driven primarily by
architectural design, ensembling, and training data, while retaining
voxel-level objectives. Examples include a 3D FC-DenseNet trained with an
asymmetric similarity loss~\cite{hashemi_asymmetric_2019}, LST-AI, an
ensemble of three 3D U-Nets trained with Tversky and cross-entropy
loss~\cite{wiltgen_lst-ai_2024}, and the nnU-Net-based
FLAMeS~\cite{dereskewicz_flames_2025}. SegHeD+ adds longitudinal,
anatomical, and volumetric constraints and lesion-aware augmentation to
train on heterogeneous datasets~\cite{basaran_seghed_2024}, and the ICPR
2024 competition on the MSLesSeg dataset benchmarked fully automated
methods across hospitals and timepoints~\cite{rondinella_icpr_2024}.
Although these methods generalize well across centers, small lesions remain
a major source of error: most lesions missed by FLAMeS were smaller than
10~mm$^3$~\cite{dereskewicz_flames_2025}. Recent work has accordingly moved
toward the individual lesion as the unit of analysis. ConfLUNet segments
MS lesion instances, including confluent lesions, and reports gains in
lesion-level F1~\cite{wynen_conflunet_2025}, and lesion-wise detection
metrics have been advocated over the DSC because they reflect what matters
clinically, including small lesions~\cite{basit_rethinking_2026}. These
works address lesion-level evaluation and inference, whereas the training
objective of most MS segmentation models remains voxel-level.

\paragraph{Remaining gap.}
This review highlights three limitations. First, voxel-, class-, and
region-level losses do not explicitly operate on individual lesions.
Second, instance-level losses balance the contributions of lesions of
different sizes but still evaluate each lesion through overlap. They
therefore do not distinguish between detecting a lesion and delineating it
accurately. Third, although MIL directly expresses this detection
criterion, its use in segmentation has largely been confined to weakly
supervised settings. No existing objective combines size-balanced overlap
with an explicit, overlap-independent detection term for each lesion.
\Cref{sec:method} introduces such an objective by adding a
component-adaptive Tversky term for balance and a lesion-level MIL term for
detection to the standard nnU-Net loss~\cite{isensee_nnu-net_2021},
without modifying the architecture.

\section{Method}
\label{sec:method}
\subsection{Small-Lesion Segmentation as a Sparse-Signal Learning Problem}
\label{sec:problem}

Let $\Omega \subset \mathbb{Z}^{3}$ denote the voxel domain of a 3D MRI
volume. For each voxel $i \in \Omega$, let $g_i \in \{0,1\}$ be the
ground-truth label and $p_i \in [0,1]$ the foreground probability predicted
by the segmentation network $f_{\theta}$. The set of lesion voxels is
$G = \{ i \in \Omega \mid g_i = 1 \}$, and it decomposes into $K$ disjoint
connected components, each corresponding to one lesion:
\begin{equation}
G = \bigcup_{k=1}^{K} C_k,
\qquad
C_k \cap C_j = \varnothing \quad \text{for } k \neq j ,
\end{equation}
where $|C_k|$ is the number of voxels in lesion $C_k$. In multiple
sclerosis the lesion signal is sparse in two ways. Lesions occupy a very
small part of the volume, $|G| \ll |\Omega|$, and their sizes are highly
unequal: lesions with an area below 10~mm$^2$ account for nearly 20\% of
all lesions but for about 1\% of the total lesion load on
average~\cite{wang_survey_1997}.

A voxel-wise loss such as cross-entropy is a sum of per-voxel terms
$\ell(p_i, g_i)$, which can be grouped by lesion:
\begin{equation}
\mathcal{L}_{\mathrm{vox}}
=
\sum_{i \in \Omega} \ell(p_i, g_i)
=
\sum_{k=1}^{K} \sum_{i \in C_k} \ell(p_i, 1)
+
\sum_{i \notin G} \ell(p_i, 0) .
\label{eq:sl_voxel_loss}
\end{equation}
Lesion $C_k$ enters the loss, and therefore the gradient, through $|C_k|$
terms, so its influence on training is proportional to its size.
Background and large lesions dominate, and a lesion of a few voxels barely
changes the objective~\cite{kofler_blob_2023}. Overlap-based losses behave
in the same way and are additionally unstable on small
objects~\cite{azad_loss_2023}. For example, if a prediction is perfect
except that one lesion $C_m$ is missed completely, the DSC is
\begin{equation}
\mathrm{DSC}
=
\frac{2\,(|G| - |C_m|)}{2|G| - |C_m|}
=
1 - \frac{|C_m|}{2|G| - |C_m|} .
\label{eq:sl_missed_lesion}
\end{equation}
A lesion that carries 1\% of the lesion volume lowers the DSC only to
about 0.995, although one of the $K$ lesions has not been detected at all.
Voxel-level overlap therefore does not control lesion-level detection.

The question addressed in this work is: \textit{how can the training
objective be redesigned so that small lesions contribute to optimization as
much as large ones, and so that every lesion is detected?} These are two
separate requirements. The first concerns \emph{balance}: the weight of
lesion $C_k$ in the loss should not scale with $|C_k|$. The second concerns
\emph{detection}: the complete omission of a lesion should be penalized
directly, at the level of the lesion rather than of its voxels. Both are
imposed on the loss only; the network $f_{\theta}$ is left unchanged, so
that differences in performance can be attributed to the objective.

\subsection{CATMIL Objective}

The proposed objective, CATMIL, adds two terms to the standard nnU-Net
loss, one for each requirement: a Component-Adaptive Tversky (CAT) term to
balance lesions and a lesion-level MIL term for detection. The connected
components $C_1,\ldots,C_K$ of the ground-truth foreground are computed
from the ground-truth mask of each training patch using 6-connectivity,
consistent with the definition of a lesion used for evaluation. They are
therefore fixed with respect to the prediction.

\subsubsection{Structure-Level Component-Adaptive Tversky Term}

According to \cref{eq:sl_voxel_loss}, a voxel-wise loss gives lesion $C_k$
an influence proportional to its size $|C_k|$. The CAT term removes this
dependence by weighting each voxel according to the component it belongs
to:
\begin{equation}
w_i =
\begin{cases}
(|C_k| + \varepsilon_{c})^{-\gamma}, & \text{if } i \in C_k, \\
w_{\mathrm{bg}}, & \text{if } g_i = 0 .
\end{cases}
\label{eq:sl_cat_weights}
\end{equation}
The exponent $\gamma \geq 0$ controls the degree of size compensation.
When $\gamma=0$, the weighting remains proportional to the number of
voxels. With $\gamma=1$, as used in this work, the total weight of lesion
$C_k$ is
\begin{equation}
\rho_k
=
\sum_{i \in C_k} w_i
=
\frac{|C_k|}{|C_k| + \varepsilon_{c}} ,
\label{eq:sl_cat_lesion_weight}
\end{equation}
which is bounded by one and no longer increases linearly with lesion
volume. The offset $\varepsilon_{c}=5$ voxels limits the weight assigned to
each voxel, preventing lesions consisting of only a few voxels from
receiving excessively large per-voxel weights. Thus, $\rho_k$ is $0.67$ for
a 10-voxel lesion and $0.95$ for a 100-voxel lesion. Since
$\rho_k \in [1/6, 1)$ for any lesion of at least one voxel, lesion volumes
may differ by orders of magnitude while their total weights differ by less
than a factor of six. Because the foreground weights are much less than
one, the background weight is set to $w_{\mathrm{bg}}=0.01$ so that the
false-positive term remains on a comparable scale.

With these weights, the true-positive, false-positive, and false-negative
sums are
\begin{equation}
\mathrm{TP}_{w} = \sum_{i \in \Omega} w_i\, p_i\, g_i,
\qquad
\mathrm{FP}_{w} = \sum_{i \in \Omega} w_i\, p_i\, (1-g_i),
\qquad
\mathrm{FN}_{w} = \sum_{i \in \Omega} w_i\, (1-p_i)\, g_i ,
\end{equation}
and the CAT term is one minus the corresponding Tversky
index~\cite{salehi_tversky_2017}:
\begin{equation}
\mathcal{L}_{\mathrm{CAT}}
=
1 -
\frac{\mathrm{TP}_{w}+\delta}
{\mathrm{TP}_{w} + \alpha\, \mathrm{FP}_{w} + \beta\, \mathrm{FN}_{w} + \delta},
\label{eq:sl_cat_loss}
\end{equation}
where $\alpha=0.3$ and $\beta=0.7$ weight false positives and false
negatives, respectively, as in the Tversky baseline. The parameter
$\delta=10^{-5}$ is the usual smoothing constant. The term is set to zero
when the training batch contains no lesions.

Under these weights, the sums take a simple form. Let
$r_k = \frac{1}{|C_k|}\sum_{i \in C_k} p_i \in [0,1]$ be the mean predicted
probability inside lesion $C_k$, that is, its soft recall. Then
\begin{equation}
\mathrm{TP}_{w} = \sum_{k=1}^{K} \rho_k\, r_k,
\qquad
\mathrm{FN}_{w} = \sum_{k=1}^{K} \rho_k\, (1-r_k) .
\label{eq:sl_cat_lesion_form}
\end{equation}
Lesions therefore contribute to the loss through their recalls $r_k$, with
coefficients $\rho_k$ that are nearly independent of lesion size. A
completely missed lesion ($r_m=0$) transfers a total weight of
$\rho_m < 1$ from $\mathrm{TP}_{w}$ to $\mathrm{FN}_{w}$, whereas in
\cref{eq:sl_missed_lesion} its effect on the DSC is proportional to
$|C_m|$. The gradient is balanced in the same way: for a voxel
$j \in C_k$,
$\partial \mathcal{L}_{\mathrm{CAT}} / \partial p_j = -c\, w_j$ with a factor
$c>0$ common to all foreground voxels (\cref{appendix:catmil-properties}),
so the total gradient magnitude received by lesion $C_k$ is $c\,\rho_k$
instead of increasing with $|C_k|$. For $\gamma=0$ and $w_{\mathrm{bg}}=1$,
the term reduces to the standard Tversky loss; if additionally
$\alpha=\beta=0.5$, it is the soft Dice loss. The CAT term can therefore be
viewed as a Tversky loss in which the lesion, rather than the voxel, is the
effective unit of counting.

\subsubsection{Lesion-Level Multiple Instance Learning Term}

The CAT term balances lesions but still evaluates them by overlap: it is
linear in $r_k$ and does not distinguish between a partially segmented
lesion and one that is not detected at all. The second requirement is
addressed using the Multiple Instance Learning
paradigm~\cite{ilse_attention-based_2018}. Each lesion is treated as a bag
of voxels. A bag is considered detected if at least one of its voxels
receives a high foreground probability. The detection score of lesion
$C_k$ is therefore its strongest response,
\begin{equation}
s_k = \max_{i \in C_k} p_i ,
\end{equation}
and the MIL term is the average negative log-score over lesions:
\begin{equation}
\mathcal{L}_{\mathrm{MIL}}
=
-\frac{1}{K} \sum_{k=1}^{K} \log \max(s_k, \varepsilon) ,
\label{eq:sl_mil_loss}
\end{equation}
where $\varepsilon=10^{-6}$ keeps each contribution finite and $K$ denotes
the number of lesions in the training batch. If the batch contains no
lesions, the term is set to zero.

This term directly penalizes the complete omission of a lesion. If lesion
$C_k$ is missed, all $p_i$ in $C_k$ are close to zero, so $s_k \approx 0$
and its contribution, $-\log s_k$, is large. Once a single voxel inside the
lesion has a high predicted probability, $s_k$ approaches one and the
contribution vanishes. The gradient acts on the maximal voxel of each
lesion with magnitude $1/(K s_k)$ (\cref{appendix:catmil-properties}).
It does not depend on $|C_k|$ and is largest for lesions that are currently
missed. Consequently, a lesion consisting of only a few voxels receives the
same detection signal as a large lesion. The term does not require the
entire lesion to be segmented; delineation is left to the overlap terms.

\subsubsection{Objective Function}

The complete objective adds the two terms to the default nnU-Net loss. In
this binary setting, the default loss is the sum of the Dice and
cross-entropy losses,
$\mathcal{L}_{\mathrm{base}} = \mathcal{L}_{\mathrm{Dice}} + \mathcal{L}_{\mathrm{CE}}$:
\begin{equation}
\mathcal{L}_{\mathrm{CATMIL}}
=
\mathcal{L}_{\mathrm{base}}
+
\lambda_{\mathrm{CAT}}(t)\, \mathcal{L}_{\mathrm{CAT}}
+
\lambda_{\mathrm{MIL}}\, \mathcal{L}_{\mathrm{MIL}} .
\label{eq:sl_catmil_loss}
\end{equation}
The base loss provides voxel-level classification and global overlap, the
CAT term provides component-balanced overlap, and the MIL term provides
lesion-level detection. Only the loss is modified; the network $f_{\theta}$
and, therefore, its hypothesis class remain unchanged. Both auxiliary terms
are bounded and continuous in the predictions
(\cref{appendix:catmil-properties}), so their influence is controlled by
the two coefficients. The base loss retains nnU-Net's deep supervision,
whereas the CAT and MIL terms are computed only on the full-resolution
output.

\paragraph{Training strategy.}
The MIL term is assigned a constant weight of
$\lambda_{\mathrm{MIL}}=0.2$. The CAT term is introduced using a linear
warm-up:
\begin{equation}
\lambda_{\mathrm{CAT}}(t)
=
\lambda_{\mathrm{CAT}} \cdot \min\left(\frac{t}{T},\, 1\right),
\qquad
\lambda_{\mathrm{CAT}} = 0.1 ,
\end{equation}
where $t$ is the current training epoch and $T=50$ epochs is the warm-up
duration. Because the CAT term changes how lesions contribute to the loss,
applying it at full strength from the first epoch may destabilize
optimization before the model has learned basic foreground-background
separation. The warm-up allows the model to first establish stable
voxel-level segmentation and then gradually shifts the supervision toward
component balance. The weights $\lambda_{\mathrm{CAT}}$ and
$\lambda_{\mathrm{MIL}}$ were fixed before the experiments and are used
unchanged for both datasets; their effect is examined in a sensitivity
analysis in \cref{appendix:catmil-loss-weights}.

\section{Experiments}
\label{sec:experiments}
\subsection{Datasets}

\paragraph{MSLesSeg.}
The primary experiments use the MSLesSeg
dataset~\cite{guarnera_mslesseg_2025} of longitudinal brain MRI scans of
patients with multiple sclerosis. Of its 75 subjects, the 53 with manual
lesion annotations are used here, as the official test set is unlabeled.
Each subject has one to four timepoints, each with T1-weighted (T1-w),
T2-weighted (T2-w), and Fluid-Attenuated Inversion Recovery (FLAIR)
images. The dataset organizers provide the images skull-stripped,
co-registered to the MNI152 template, and resampled to 1~mm isotropic
spacing.

\paragraph{3D-MR-MS.}
To assess whether the behavior of CATMIL carries over to other data, the
five losses are also trained and evaluated on a second multiple sclerosis
dataset, 3D-MR-MS~\cite{lesjak_novel_2018}, which was acquired at a
different center with different imaging conditions. It contains 30
patients with consensus-based white matter lesion annotations, provided
co-registered and bias-field corrected. Of the available modalities, only
T1-w, T2-w, and FLAIR are used, to keep the input configuration identical
to MSLesSeg. The voxels of this dataset are anisotropic
($0.47 \times 0.47 \times 0.8$~mm), so lesion sizes on 3D-MR-MS are given
in mm$^3$ rather than in voxels.

\subsection{Data Split}

Data are partitioned at the patient level, so that scans of one patient
never appear in both training and evaluation sets. For MSLesSeg, six
patients, with all their timepoints (12 test cases), are held out as an
independent test set; for 3D-MR-MS, six of the 30 patients are held out.
The remaining patients (47 and 24, respectively) are divided into five
folds following the nnU-Net cross-validation strategy. Each of the five
fold models is evaluated on the test set, and results are reported as the
mean and sample standard deviation across the five models; statistical
comparisons are described in \cref{sec:stats}. All training
settings, including the CATMIL loss weights, are identical for the two
datasets, so 3D-MR-MS was not used to tune the objective.

\subsection{Compared Losses}

Five configurations of nnU-Net are compared, differing only in the loss
function:
\begin{itemize}
    \item \textbf{DiceCE}: the default Dice and cross-entropy loss;
    \item \textbf{Tversky}~\cite{salehi_tversky_2017}: Tversky loss, with
    $\alpha=0.3$ and $\beta=0.7$ weighting false positives and false
    negatives;
    \item \textbf{FocalTversky}~\cite{abraham_novel_2019}: Focal Tversky
    loss, $\mathcal{L}_{\mathrm{Tversky}}^{1/\gamma_{\mathrm{F}}}$, with the
    same $\alpha$ and $\beta$ and focal exponent $\gamma_{\mathrm{F}}=4/3$;
    \item \textbf{CC-DiceCE}~\cite{bouteille_learning_2026}: an
    instance-level loss that adds a per-lesion DiceCE term, built on the
    component-wise evaluation protocol of CC-Metrics~\cite{jaus_every_2025},
    to the global DiceCE loss with equal weight;
    \item \textbf{CATMIL}: the proposed loss.
\end{itemize}
Tversky and FocalTversky replace the DiceCE loss, whereas CC-DiceCE and
CATMIL add terms to it. CC-DiceCE is the closest prior objective: like the
CAT term, it balances lesions of different sizes, but it assesses each
lesion through overlap and has no detection term. It is the authors'
published loss, ported unchanged into the same nnU-Net pipeline and
150-epoch budget rather than trained with the authors' own setup. Ablations of the two CATMIL terms and of their weights are
reported in \cref{sec:ablation,appendix:catmil-loss-weights}.

\subsection{Implementation}

All models use nnU-Net (version 2.1.1)~\cite{isensee_nnu-net_2021} with
its default preprocessing, patch-based training, and optimizer: stochastic
gradient descent with Nesterov momentum of 0.99, weight decay of
$3\times10^{-5}$, and an initial learning rate of 0.01 with polynomial
decay. Training runs for 150 epochs with a batch size of 2 on an NVIDIA
Quadro RTX 8000 GPU. The loss implementation and evaluation code are
available at \url{https://github.com/luumsk/SmallLesionMRI}.

\subsection{Evaluation Metrics}

Three groups of metrics are used: voxel-level metrics for overall
segmentation quality, lesion-level metrics for detection, in particular of
small lesions, and error-volume metrics for the extent of false-positive
and false-negative errors. Lesion-level metrics are included because the
DSC is dominated by large lesions (\cref{eq:sl_missed_lesion}) and does not
reflect whether individual, clinically relevant lesions are
found~\cite{basit_rethinking_2026,jaus_every_2025}.

\paragraph{Voxel-level metrics.}
Overall segmentation quality is measured with the DSC~\cite{dice_measures_1945}
and the 95th-percentile Hausdorff distance (HD95, in
mm)~\cite{taha_metrics_2015}. The name Dice is kept only for the Dice loss
and the DiceCE baseline, which use the differentiable form of the
coefficient.

\paragraph{Lesion definition and detection.}
A lesion is a 3D connected component of the binary mask under
6-connectivity, the same definition as in the proposed loss. A reference
lesion counts as detected if at least one predicted voxel overlaps it, and
a predicted component counts as correct if it overlaps at least one
reference lesion. This any-overlap criterion is lenient: a prediction that
covers only part of a lesion still counts as a detection. To test whether
the results depend on it, lesion recall was also computed with stricter
criteria that require at least 10\%, 25\%, or 50\% of the lesion voxels to
be covered (\cref{appendix:strict-detection}).

\paragraph{Lesion-level metrics.}
Lesion-wise recall is the fraction of reference lesions that are detected,
lesion-wise precision is the fraction of predicted components that are
correct, and the lesion-wise F1 score combines the two. Because precision
counts components regardless of their size, many small false-positive
components lower it substantially even when their total volume is small.
Small Lesion Recall is the lesion-wise recall restricted to reference
lesions of at most 150 voxels, which corresponds to 150~mm$^3$ at the 1~mm
isotropic spacing of MSLesSeg and to about 26~mm$^3$ on 3D-MR-MS. The FN
Count is the number of reference lesions that are completely missed. The
recall of lesions of at least 3~mm is the lesion-wise recall restricted to
lesions above the size used in clinical MRI reading: current consensus
recommendations state that multiple sclerosis lesions should generally be
at least 3~mm in longest diameter, with exceptions for lesions in the
corpus callosum, the floor of the fourth ventricle, and the outer margin
of the pons~\cite{barkhof_2024_2025,filippi_mri_2016}. The longest diameter
of a reference lesion is measured as its largest in-plane extent over all
axial, coronal, and sagittal slices.

\paragraph{Error-volume metrics.}
The FP Volume is the physical volume of the false-positive voxels. The FP
Blob Fraction is the share of this volume that lies in false-positive
components more than 2~mm from any reference lesion, separating detached
false alarms from extensions of true lesions. The FN Volume Fraction is
the share of the total reference lesion volume that belongs to missed
lesions.

All metrics are computed per test case and averaged over cases; their
exact definitions are given in \cref{appendix:metrics}.

\subsection{Statistical Analysis}
\label{sec:stats}

The standard deviations in the result tables describe the variation
between the five fold models. They do not describe the uncertainty that
comes from the small test sets, so the differences between CATMIL and each
baseline are also tested with the patient as the independent unit. This
matters for MSLesSeg, whose 12 test cases are longitudinal scans of six
patients.

For each metric, the difference between CATMIL and a baseline is computed
in the same way as the table values: the mean over test cases of the
five-model mean. Its 95\% confidence interval (CI) is obtained with a
two-level bootstrap with 10{,}000 resamples. Each resample draws patients
with replacement, together with all their scans, and draws fold models with
replacement. The fold models are drawn jointly for both losses, because
fold $k$ of every loss is trained on the same data split. For lesion-level
recall, the same bootstrap is applied to the pooled reference lesions of
the resampled patients. A paired exact Wilcoxon signed-rank test is applied
to the per-patient differences, with the Holm correction over the four
baselines.

With six patients, the smallest possible two-sided $p$-value of the exact
test is $1/32 \approx 0.031$ before correction and $0.125$ after it, so no
comparison can reach $p < 0.05$ after correction. A difference is
therefore regarded as supported when its 95\% CI excludes zero, and the
number of patients in which CATMIL is better is reported with it. These
analyses describe the consistency of an effect across the available
patients; they are not population-level estimates. To obtain a larger
sample, the MSLesSeg comparisons are also repeated on the cross-validation
predictions of the 47 training patients (\cref{sec:catmil_oof}).

\section{Results}
\label{sec:results}
The results are organized around the question posed in the introduction.
\Cref{sec:main_results} establishes the main effect, \cref{sec:where_gain}
and \cref{sec:no_response} show where the gain arises and how the
baselines fail, \cref{sec:arch_vs_objective} and \cref{sec:ablation}
attribute the failure to the objective and separate the roles of the two
terms,
\cref{sec:fp_pp} characterizes the cost, and \cref{sec:robustness} tests
whether the findings hold beyond the six MSLesSeg test patients.

\subsection{CATMIL Detects More Small Lesions at Comparable Overlap}
\label{sec:main_results}

\begin{table}[t]
\centering
\caption{Comparison of CATMIL with standard losses and the instance-level
loss CC-DiceCE on the two test sets, without post-processing. Values are
the mean $\pm$ sample standard deviation over the five fold models. Best
values are shown in \textbf{bold}; second-best values are
\underline{underlined}. Paired differences with patient-level CIs are given
in \cref{tab:paired_differences}.}
\label{tab:loss_comparison_small_lesion}
\resizebox{\textwidth}{!}{%
\begin{tabular}{lccccc}
\toprule
\textbf{Metric}
& \textbf{CATMIL (ours)}
& \textbf{DiceCE}
& \textbf{Tversky}
& \textbf{FocalTversky}
& \textbf{CC-DiceCE} \\
\midrule
\multicolumn{6}{l}{\emph{MSLesSeg}} \\
Small Lesion Recall $\uparrow$
& \textbf{0.873 $\pm$ 0.01}
& 0.796 $\pm$ 0.03
& \underline{0.834 $\pm$ 0.02}
& 0.831 $\pm$ 0.02
& 0.831 $\pm$ 0.01 \\
FN Count $\downarrow$
& \textbf{2.57 $\pm$ 0.15}
& 4.95 $\pm$ 0.40
& 3.77 $\pm$ 0.23
& 3.70 $\pm$ 0.24
& \underline{3.65 $\pm$ 0.33} \\
FN Volume Fraction $\downarrow$
& \textbf{0.021 $\pm$ 0.003}
& 0.034 $\pm$ 0.01
& 0.026 $\pm$ 0.01
& \underline{0.025 $\pm$ 0.005}
& 0.035 $\pm$ 0.01 \\
DSC $\uparrow$
& \textbf{0.783 $\pm$ 0.01}
& \underline{0.780 $\pm$ 0.01}
& 0.771 $\pm$ 0.01
& \underline{0.780 $\pm$ 0.01}
& 0.778 $\pm$ 0.01 \\
HD95 (mm) $\downarrow$
& \textbf{7.98 $\pm$ 1.84}
& 9.04 $\pm$ 1.34
& 10.24 $\pm$ 0.15
& \underline{8.21 $\pm$ 1.93}
& 9.72 $\pm$ 1.53 \\
Lesion F1 $\uparrow$
& 0.757 $\pm$ 0.02
& \underline{0.843 $\pm$ 0.01}
& 0.840 $\pm$ 0.01
& \textbf{0.846 $\pm$ 0.01}
& 0.807 $\pm$ 0.01 \\
FP Volume (mm$^3$) $\downarrow$
& \underline{1537 $\pm$ 245}
& 1819 $\pm$ 654
& 2621 $\pm$ 559
& 2282 $\pm$ 710
& \textbf{1514 $\pm$ 289} \\
\midrule
\multicolumn{6}{l}{\emph{3D-MR-MS}} \\
Small Lesion Recall $\uparrow$
& \textbf{0.618 $\pm$ 0.03}
& 0.537 $\pm$ 0.01
& 0.564 $\pm$ 0.02
& 0.563 $\pm$ 0.02
& \underline{0.579 $\pm$ 0.01} \\
FN Count $\downarrow$
& \textbf{16.9 $\pm$ 0.64}
& 22.0 $\pm$ 0.94
& 21.3 $\pm$ 1.15
& 21.3 $\pm$ 1.20
& \underline{19.1 $\pm$ 1.06} \\
FN Volume Fraction $\downarrow$
& 0.085 $\pm$ 0.03
& \textbf{0.068 $\pm$ 0.01}
& 0.082 $\pm$ 0.03
& \underline{0.074 $\pm$ 0.03}
& 0.113 $\pm$ 0.03 \\
DSC $\uparrow$
& 0.718 $\pm$ 0.01
& \underline{0.730 $\pm$ 0.02}
& 0.726 $\pm$ 0.01
& \textbf{0.734 $\pm$ 0.01}
& 0.720 $\pm$ 0.02 \\
HD95 (mm) $\downarrow$
& \textbf{11.55 $\pm$ 1.27}
& \underline{11.94 $\pm$ 1.79}
& 13.38 $\pm$ 3.65
& 12.27 $\pm$ 3.10
& 13.91 $\pm$ 3.05 \\
Lesion F1 $\uparrow$
& 0.471 $\pm$ 0.04
& \textbf{0.668 $\pm$ 0.02}
& \underline{0.640 $\pm$ 0.02}
& 0.637 $\pm$ 0.03
& 0.602 $\pm$ 0.03 \\
FP Volume (mm$^3$) $\downarrow$
& \textbf{1232 $\pm$ 165}
& \underline{1234 $\pm$ 213}
& 1994 $\pm$ 284
& 1958 $\pm$ 269
& 1283 $\pm$ 189 \\
\bottomrule
\end{tabular}%
}
\end{table}

\begin{table}[t]
\centering
\caption{Paired differences between CATMIL and each baseline (CATMIL minus
baseline), with 95\% patient-level bootstrap CIs and, in parentheses, the
number of test patients in which CATMIL is better. Differences whose CI
excludes zero are shown in \textbf{bold}. PP5: after removing predicted
components smaller than 5 voxels; DSC, HD95, and FP Volume change by less
than 0.002, 0.15~mm, and 25~mm$^3$ under PP5 and are not repeated. With
six patients, the smallest possible Holm-corrected Wilcoxon $p$-value is
0.125 (\cref{sec:stats}).}
\label{tab:paired_differences}
\resizebox{\textwidth}{!}{%
\begin{tabular}{lllll}
\toprule
\textbf{Metric} & \textbf{vs.\ DiceCE} & \textbf{vs.\ Tversky} & \textbf{vs.\ FocalTversky} & \textbf{vs.\ CC-DiceCE} \\
\midrule
\multicolumn{5}{l}{\emph{MSLesSeg}} \\
Small Lesion Recall $\uparrow$ & \textbf{+0.077 [+0.030, +0.157] (6/6)} & \textbf{+0.039 [+0.004, +0.093] (6/6)} & \textbf{+0.042 [+0.005, +0.101] (6/6)} & \textbf{+0.042 [+0.020, +0.071] (6/6)} \\
FN Count $\downarrow$ & \textbf{$-$2.38 [$-$5.29, $-$0.84] (6/6)} & \textbf{$-$1.20 [$-$2.98, $-$0.27] (6/6)} & \textbf{$-$1.13 [$-$2.71, $-$0.27] (5/6)} & \textbf{$-$1.08 [$-$1.95, $-$0.53] (6/6)} \\
DSC $\uparrow$ & +0.004 [$-$0.010, +0.031] (1/6) & \textbf{+0.013 [+0.001, +0.034] (5/6)} & +0.003 [$-$0.015, +0.023] (5/6) & +0.006 [$-$0.004, +0.017] (6/6) \\
HD95 (mm) $\downarrow$ & $-$1.06 [$-$4.67, +0.73] (4/6) & $-$2.26 [$-$7.35, +0.11] (4/6) & $-$0.22 [$-$2.92, +2.72] (4/6) & \textbf{$-$1.73 [$-$5.43, $-$0.01] (5/6)} \\
Lesion F1 $\uparrow$ & \textbf{$-$0.086 [$-$0.138, $-$0.043] (0/6)} & \textbf{$-$0.083 [$-$0.143, $-$0.038] (0/6)} & \textbf{$-$0.088 [$-$0.151, $-$0.039] (0/6)} & \textbf{$-$0.049 [$-$0.091, $-$0.018] (0/6)} \\
FP Volume (mm$^3$) $\downarrow$ & $-$282 [$-$1582, +271] (1/6) & \textbf{$-$1084 [$-$2675, $-$302] (6/6)} & \textbf{$-$746 [$-$2189, $-$182] (6/6)} & +23 [$-$276, +236] (1/6) \\
\midrule
\multicolumn{5}{l}{\emph{MSLesSeg, PP5}} \\
Small Lesion Recall $\uparrow$ & \textbf{+0.070 [+0.023, +0.151] (6/6)} & \textbf{+0.033 [+0.001, +0.079] (6/6)} & \textbf{+0.036 [+0.000, +0.093] (6/6)} & \textbf{+0.041 [+0.017, +0.073] (6/6)} \\
FN Count $\downarrow$ & \textbf{$-$2.10 [$-$4.80, $-$0.66] (6/6)} & \textbf{$-$0.93 [$-$2.38, $-$0.13] (6/6)} & \textbf{$-$0.95 [$-$2.40, $-$0.14] (6/6)} & \textbf{$-$1.05 [$-$1.84, $-$0.46] (6/6)} \\
Lesion F1 $\uparrow$ & $-$0.017 [$-$0.040, +0.001] (2/6) & $-$0.012 [$-$0.038, +0.006] (2/6) & $-$0.017 [$-$0.043, +0.001] (2/6) & +0.003 [$-$0.010, +0.017] (4/6) \\
\midrule
\multicolumn{5}{l}{\emph{3D-MR-MS}} \\
Small Lesion Recall $\uparrow$ & \textbf{+0.081 [+0.034, +0.127] (5/6)} & \textbf{+0.054 [+0.019, +0.086] (5/6)} & \textbf{+0.055 [+0.021, +0.087] (5/6)} & +0.039 [$-$0.001, +0.080] (4/6) \\
FN Count $\downarrow$ & \textbf{$-$5.13 [$-$12.40, $-$0.50] (5/6)} & \textbf{$-$4.43 [$-$11.23, $-$0.27] (5/6)} & \textbf{$-$4.43 [$-$11.47, $-$0.23] (5/6)} & \textbf{$-$2.27 [$-$4.80, $-$0.47] (5/6)} \\
DSC $\uparrow$ & $-$0.012 [$-$0.032, +0.010] (0/6) & $-$0.008 [$-$0.030, +0.013] (3/6) & $-$0.016 [$-$0.035, +0.001] (1/6) & $-$0.002 [$-$0.015, +0.013] (2/6) \\
HD95 (mm) $\downarrow$ & $-$0.38 [$-$4.14, +2.07] (3/6) & $-$1.83 [$-$8.51, +1.50] (2/6) & $-$0.72 [$-$6.07, +2.92] (3/6) & $-$2.36 [$-$8.87, +0.60] (4/6) \\
Lesion F1 $\uparrow$ & \textbf{$-$0.197 [$-$0.258, $-$0.142] (0/6)} & \textbf{$-$0.168 [$-$0.242, $-$0.106] (0/6)} & \textbf{$-$0.166 [$-$0.232, $-$0.108] (0/6)} & \textbf{$-$0.131 [$-$0.195, $-$0.066] (0/6)} \\
FP Volume (mm$^3$) $\downarrow$ & $-$2 [$-$110, +110] (3/6) & \textbf{$-$761 [$-$1587, $-$118] (6/6)} & \textbf{$-$726 [$-$1550, $-$92] (6/6)} & $-$50 [$-$145, +16] (3/6) \\
\bottomrule
\end{tabular}%
}
\end{table}

\Cref{tab:loss_comparison_small_lesion,tab:paired_differences} give the
main comparison. Three observations frame the rest of the section.

\textbf{CATMIL detects more small lesions than every compared loss.} On
MSLesSeg, it raises small-lesion recall from 0.796 with DiceCE to 0.873,
and misses 48\% fewer lesions per case. The gain is present in all six test
patients against every baseline, with CIs that exclude zero
(\cref{tab:paired_differences}). The three alternatives that already
emphasize false negatives or individual lesions, Tversky, FocalTversky, and
CC-DiceCE, all reach a similar intermediate level (0.831--0.834). The
gain does not rely on the lenient any-overlap detection criterion: when at
least half of a lesion must be covered, the gain over DiceCE remains about
8 percentage points on MSLesSeg and 10 on 3D-MR-MS
(\cref{appendix:strict-detection}).

\textbf{The recovered lesions do not come at the cost of overlap or of
lesion volume.} The DSC and HD95 of CATMIL are within the variation between
patients of every baseline on MSLesSeg, and its false-positive volume is no
larger than that of any baseline. CATMIL therefore does not find more
lesions by predicting more lesion tissue. Equally, the DSC barely registers
the recovered lesions, which is the gap between voxel- and lesion-level
evaluation that motivates this work (\cref{eq:sl_missed_lesion}).

\textbf{The cost appears in lesion-wise F1.} CATMIL has the lowest
lesion-wise F1 in every test patient. Together with its unchanged
false-positive volume, this implies many small false-positive components
rather than over-segmentation, which \cref{sec:fp_pp} confirms.

\subsection{The Gain Is a Net Gain, Concentrated Where the Baselines Fail}
\label{sec:where_gain}

\begin{figure}[t]
    \centering
    \includegraphics[width=0.85\textwidth]{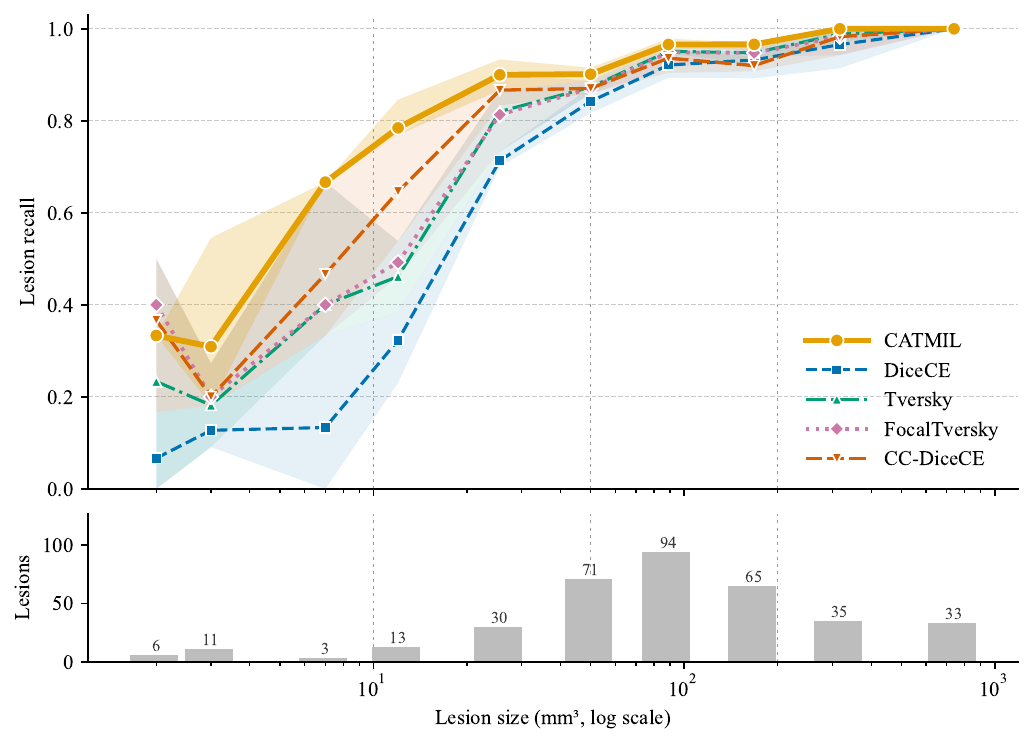}
    \caption{Lesion recall against lesion size on the MSLesSeg test set.
    Upper panel: fraction of reference lesions detected in each size
    interval (logarithmic axis); lines are the mean over the five fold
    models and shaded bands span the fold minimum to maximum. Lower panel:
    number of reference lesions per interval. Dotted vertical lines mark
    10, 50, and 200~mm$^3$. A lesion counts as detected if any predicted
    voxel overlaps it.}
    \label{fig:recall_vs_size_curve}
\end{figure}

\textbf{CATMIL moves the detection limit toward smaller lesions.}
\Cref{fig:recall_vs_size_curve} shows that the gap between CATMIL and the
baselines is widest exactly where the baselines fail, between about 5 and
50~mm$^3$, and closes above about 100~mm$^3$, where every loss detects
nearly all lesions. One way to summarize the shift is the size at which
recall reaches 0.8: about 40~mm$^3$ for DiceCE, 20--25~mm$^3$ for the
Tversky-based losses and CC-DiceCE, and about 13~mm$^3$ for CATMIL. Lesions
of at most 5~mm$^3$ remain difficult for every loss, so the size bias of
voxel-wise training is reduced but not removed.

\begin{figure}[t]
    \centering
    \includegraphics[width=\textwidth]{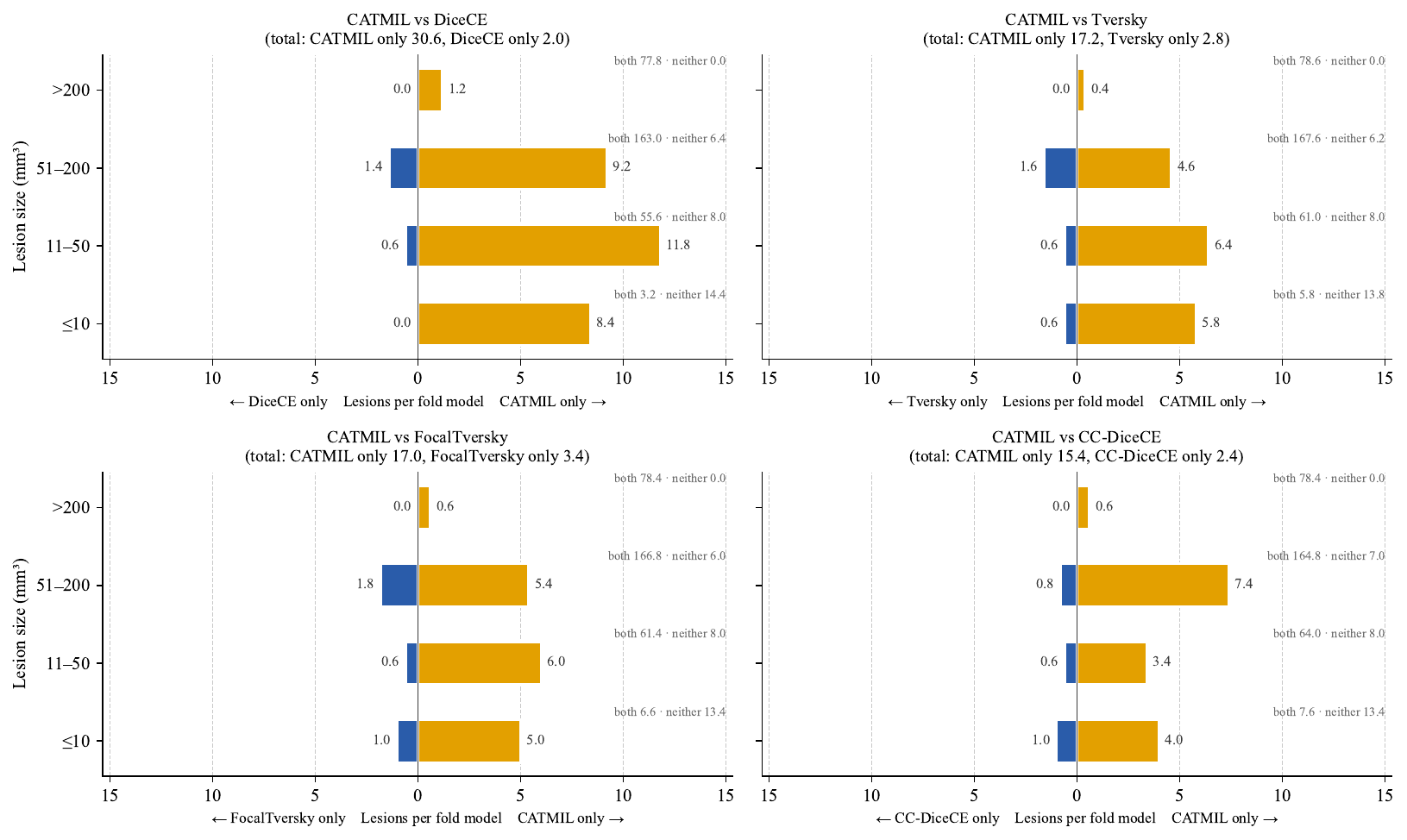}
    \caption{Lesions detected by only one of two losses on the MSLesSeg
    test set, by lesion size. Each panel pairs every reference lesion
    between CATMIL and one baseline trained on the same fold. Bars to the
    right: lesions detected by CATMIL but not by the baseline; bars to the
    left: lesions detected by the baseline but not by CATMIL. Grey text:
    lesions detected by both or by neither. Values are means over the five
    fold models.}
    \label{fig:lesion_fate}
\end{figure}

\begin{figure}[t]
    \centering
    \includegraphics[width=\textwidth]{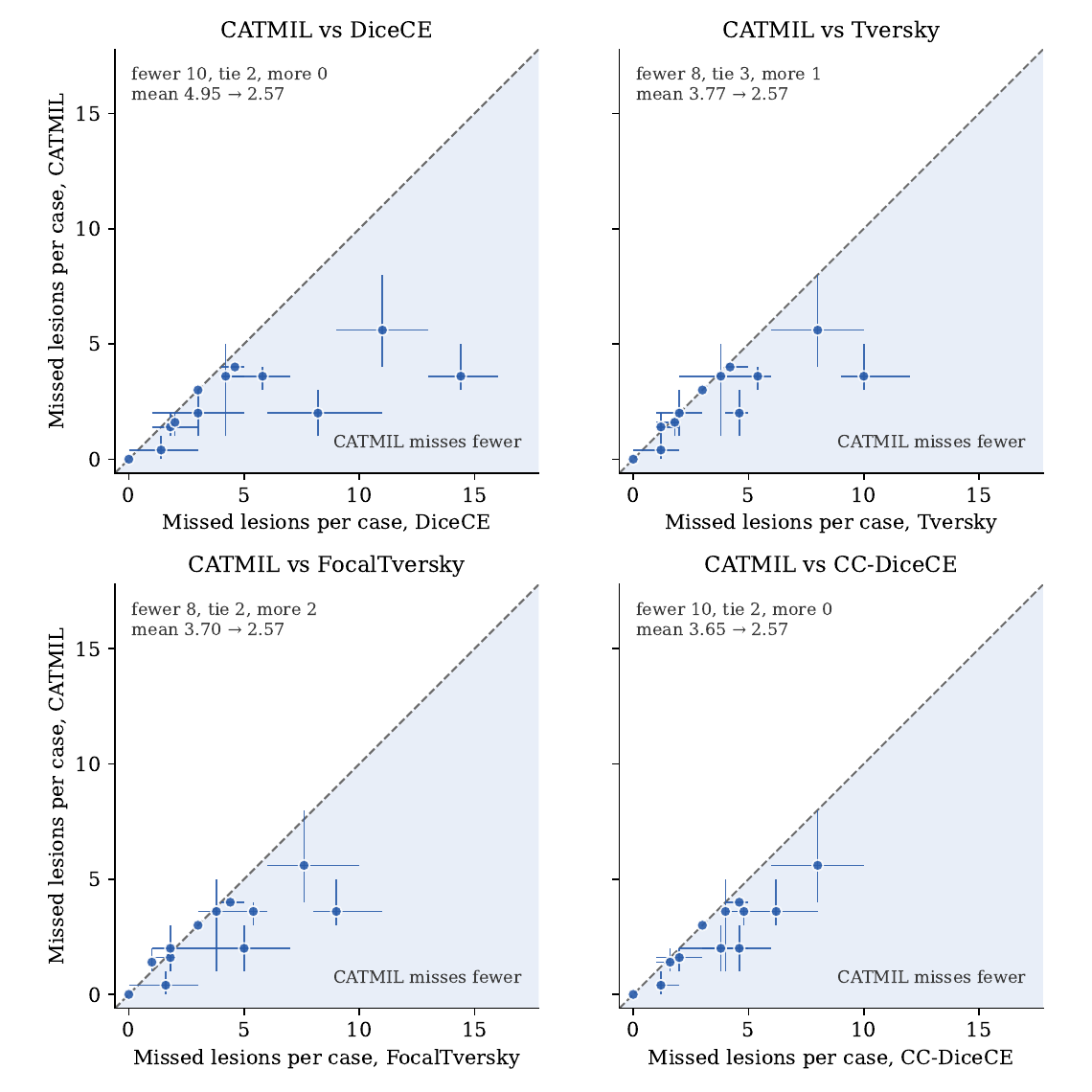}
    \caption{Missed lesions per test case for CATMIL compared with DiceCE,
    Tversky, FocalTversky, and CC-DiceCE. Each point is one test case, placed at the
    mean FN lesion count over the five fold models; error bars span the
    fold minimum to maximum. Points below the dashed diagonal (shaded
    region) are cases in which CATMIL misses fewer lesions than the
    baseline. The text in each panel gives the number of cases in which
    CATMIL misses fewer, the same number, or more lesions, and the mean
    number of missed lesions per case for the baseline and for CATMIL.}
    \label{fig:fn_per_case_paired}
\end{figure}

\textbf{The gain is not a redistribution of errors.} A higher recall could
hide a trade in which CATMIL gains some lesions but loses others.
\Cref{fig:lesion_fate} pairs every reference lesion between CATMIL and each
baseline trained on the same fold, and the exchange is strongly one-sided:
per fold model, CATMIL detects 30.6 lesions that DiceCE misses, while
DiceCE detects 2.0 that CATMIL misses, and the ratio against CC-DiceCE is
15.4 to 2.4. No lesion above 200~mm$^3$ is lost. The same holds across
cases (\cref{fig:fn_per_case_paired}): against DiceCE and CC-DiceCE, CATMIL
misses fewer lesions in 10 of the 12 test cases and more in none.

\textbf{The gain is largest in patients with many small lesions, and the
DSC does not see it.} Most of the recovered lesions come from three
high-burden cases with many small lesions
(\cref{tab:test_case_lesion_profile}). In one of them, P47\_T1, CATMIL
recovers about 11 lesions relative to DiceCE while the case-level DSC stays
at 0.85: the recovered lesions are too small to move a volume-overlap
score. This is the failure mode that \cref{eq:sl_missed_lesion} predicts,
observed on real data.

\subsection{The Baselines Do Not Respond to the Lesions They Miss}
\label{sec:no_response}

\begin{figure}[t]
    \centering
    \includegraphics[width=0.75\textwidth]{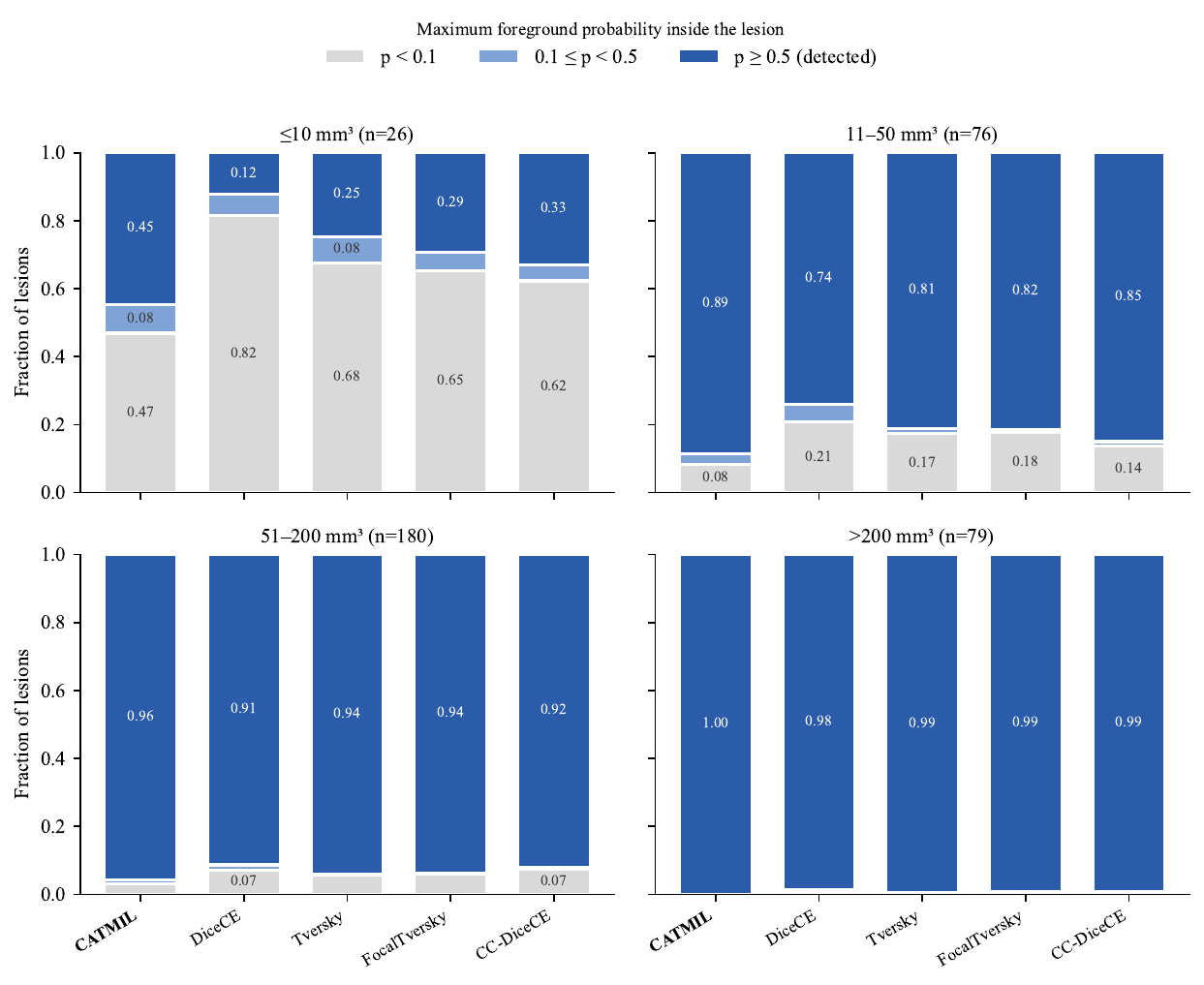}
    \caption{Maximum foreground probability inside each reference lesion on
    the MSLesSeg test set, by lesion size. Bars give the fraction of
    lesions whose maximum probability is below 0.1 (no response), between
    0.1 and 0.5 (weak response below the threshold), or at least 0.5
    (detected). Each lesion contributes once per fold model; $n$ is the
    number of reference lesions in the bin.}
    \label{fig:max_prob_by_size}
\end{figure}

A missed lesion can be missed narrowly, with a response just below the
decision threshold, or entirely, with no response at all. The distinction
matters: the first can be corrected at inference, the second only during
training. \Cref{fig:max_prob_by_size} assigns the maximum foreground
probability inside every reference lesion to one of three levels.

\textbf{The baselines miss small lesions entirely.} For lesions of at most
10~mm$^3$, DiceCE assigns a maximum probability below 0.1 to 82\% of them,
and even the best baseline, CC-DiceCE, to 62\%. The intermediate level,
a weak response below the threshold, never exceeds 8\% for any loss. A
lower threshold therefore cannot recover these lesions; the omission is
already present in the network output. CATMIL reduces the fraction with no
response to 47\%. It does not merely push weak responses over the
threshold: it makes the network respond to lesions that the other
objectives leave invisible. The same pattern holds on 3D-MR-MS
(\cref{fig:max_prob_by_size_3dmrms}).

\begin{figure}[!htbp]
    \centering
    \includegraphics[width=\textwidth]{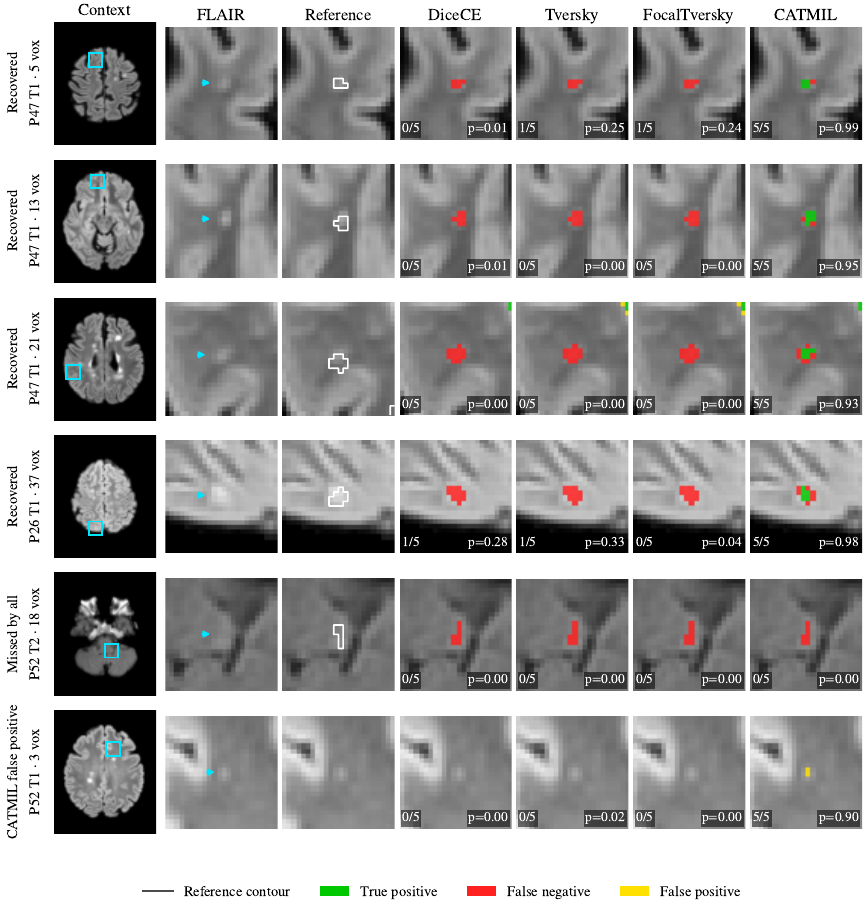}
    \caption{Lesion-centered comparison of the four losses on the MSLesSeg
    test set. Each row shows one lesion (or one CATMIL false-positive
    component), selected by the rule described in the text. From left to
    right: the full axial slice with the crop location (cyan box); a 24~mm
    FLAIR crop with the target marked by a cyan arrow; the reference
    contour (white); and the predictions of the fold-0 model of each loss.
    Green indicates correctly segmented lesion voxels, red indicates missed
    lesion voxels (false negatives), and yellow indicates false-positive
    voxels. The number at the bottom left of each prediction panel is the
    number of the five fold models that detect the lesion; $p$ at the
    bottom right is the maximum lesion probability inside the lesion,
    averaged over the five fold models. The models receive T1-w, T2-w, and
    FLAIR images; only FLAIR is shown.}
    \label{fig:qual_lesion_gallery}
\end{figure}

\begin{figure}[!htbp]
    \centering
    \includegraphics[width=0.8\textwidth]{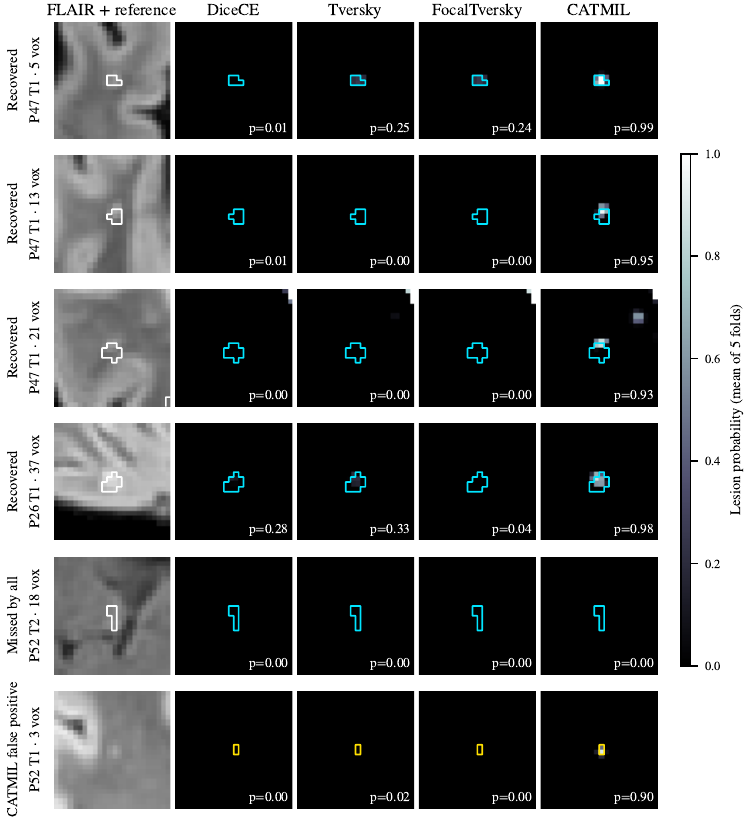}
    \caption{Lesion probability maps for the lesions in
    \cref{fig:qual_lesion_gallery}, averaged over the five fold models of
    each loss. The cyan contour marks the reference lesion and the yellow
    contour marks the CATMIL false-positive component. Values of $p$ are as
    in \cref{fig:qual_lesion_gallery}.}
    \label{fig:qual_probability_maps}
\end{figure}

\textbf{Individual lesions show the same mechanism.}
\Cref{fig:qual_lesion_gallery,fig:qual_probability_maps} show lesions
selected by a fixed rule rather than by visual inspection: a
\emph{recovered} lesion has at most 50 voxels, is detected by at least four
of the five CATMIL models, and by at most one model of each baseline; a
lesion \emph{missed by all} is detected by none of the twenty models; the
\emph{CATMIL false positive} is produced by all five CATMIL models and by no
baseline model. (The case analysis predates the CC-DiceCE experiments and
covers the other four losses.) Four observations stand out.
\begin{itemize}
\item \emph{The omission is a size effect, not a visibility effect.} The
13-voxel lesion in row~2 has median contrast and is clearly visible on
FLAIR, but it is isolated, and all fifteen baseline models assign it a
probability of at most 0.01. It accounts for 0.07\% of the lesion voxels
of its case, but for one of its 59 lesions; only a lesion-level term gives
it a share of the training signal that matches the latter.
\item \emph{CATMIL detects, but does not fully delineate.} CATMIL detects
each recovered lesion in all five models with $p \geq 0.93$, but covers
only part of it (lesion-wise DSC 0.28--0.52) and in row~1 merges a
5-voxel lesion with a neighbor 3~mm away. The detection term decides
whether a lesion is found; its extent still depends on the overlap terms.
\item \emph{Some recovered lesions are clinically informative.} The
juxtacortical lesion in row~4 lies against bright cortex and is detected
by at most one baseline model; juxtacortical location is one of the
criteria for dissemination in space in
MS~\cite{thompson_diagnosis_2018}.
\item \emph{The remaining misses and the false positive reflect the image,
not the loss.} The lesion missed by all models (row~5) has a contrast below
that of 99\% of the small test lesions, and across the test set the small
lesions that CATMIL still misses have about half the contrast of those it
detects (median 1.86 against 3.65). The false positive (row~6) is a 3-voxel
focal hyperintensity with no annotated counterpart at follow-up: the same
response to compact bright foci that recovers rows~1--4, at a location the
reference does not mark.
\end{itemize}
Lesion characteristics and the contrast measure are described in
\cref{appendix:test-case-lesion-profile}.

\subsection{Changing the Architecture Does Not Recover the Missed Lesions}
\label{sec:arch_vs_objective}

If the baselines' lack of response reflected the capacity or inductive bias
of nnU-Net, a different network trained with the same objective should
respond to more small lesions. Four architectures were trained with the
unchanged DiceCE objective under the same protocol: the residual
SegResNet~\cite{crimi_3d_2019}, the transformer-based
SwinUNETR~\cite{hatamizadeh_swin_2022} and
UNETR~\cite{hatamizadeh_unetr_2021}, and the state-space model U-Mamba in
its UMambaBot variant~\cite{ma_u-mamba_2024}, each with the same
cross-validation splits, 150-epoch budget, and its default hyperparameters
(\cref{tab:arch_vs_objective}).

\begin{table}[t]
\centering
\caption{Small-lesion detection by architecture and objective on the
MSLesSeg test set, without post-processing. All architectures are trained
with the DiceCE objective, except nnU-Net with CATMIL. Values are means
over the test scans and the five fold models. Best values are shown in
\textbf{bold}; second-best values are \underline{underlined}.}
\label{tab:arch_vs_objective}
\resizebox{\textwidth}{!}{%
\begin{tabular}{lcccccc}
\toprule
\textbf{Metric}
& \textbf{nnU-Net}
& \textbf{nnU-Net}
& \textbf{SegResNet}
& \textbf{UMambaBot}
& \textbf{SwinUNETR}
& \textbf{UNETR} \\
& \textbf{(CATMIL)}
& \textbf{(DiceCE)}
& \textbf{(DiceCE)}
& \textbf{(DiceCE)}
& \textbf{(DiceCE)}
& \textbf{(DiceCE)} \\
\midrule
Small Lesion Recall $\uparrow$
& \textbf{0.873} & \underline{0.796} & 0.779 & 0.778 & 0.771 & 0.715 \\
FN Count $\downarrow$
& \textbf{2.57} & 4.95 & 4.83 & \underline{4.65} & 4.92 & 6.43 \\
Lesion F1 $\uparrow$
& 0.757 & \textbf{0.843} & 0.836 & \underline{0.842} & 0.804 & 0.765 \\
\midrule
DSC $\uparrow$
& \textbf{0.783} & 0.780 & 0.761 & \underline{0.782} & 0.762 & 0.745 \\
HD95 (mm) $\downarrow$
& \textbf{7.98} & 9.04 & 10.58 & \underline{8.68} & 10.57 & 10.96 \\
\bottomrule
\end{tabular}%
}
\end{table}

\textbf{None of the architectures detected more small lesions than
nnU-Net.} Their small-lesion recall was similar to or below that of
nnU-Net with the same loss (0.715--0.779 against 0.796); only UNETR differed
measurably, and it was worse. Changing the objective instead raised recall
above every architecture in all six patients, by 0.094 to 0.158, with all
CIs excluding zero, and the gain remained after PP5.

\textbf{Aggregate overlap again hid the difference.} UMambaBot matched the
DSC of nnU-Net and had a lower HD95, yet detected no more small lesions. A
network can thus be competitive on voxel-level metrics while leaving the
same lesions undetected. Across the convolutional, transformer, and
state-space designs tested, the omission of small lesions depended more on
how supervision is allocated across lesions than on the network that
receives it. These architectures were not tuned individually, and CATMIL
was not combined with them, so this result shows that the tested changes of
architecture do not substitute for a change of objective, not that
architecture is irrelevant.

\subsection{Roles of the CAT and MIL Terms}
\label{sec:ablation}

CATMIL combines a balance term and a detection term. To separate their
roles, each was added alone to the DiceCE loss under the same protocol
(\cref{tab:ablation_cat_mil}).

\begin{table}[t]
\centering
\caption{Ablation of the two CATMIL terms on the MSLesSeg test set without
post-processing. DiceCE\,+\,CAT and DiceCE\,+\,MIL add only one of the two
terms to the base loss. Values are means over the five fold models;
$\Delta$ is the paired difference in small-lesion recall from DiceCE with
95\% patient-level CI; differences whose CI excludes zero are shown in
\textbf{bold}.}
\label{tab:ablation_cat_mil}
\resizebox{\textwidth}{!}{%
\begin{tabular}{lccccl}
\toprule
\textbf{Loss} & \textbf{Small Recall} & \textbf{Lesion F1} & \textbf{DSC} & \textbf{HD95 (mm)} & $\Delta$ \textbf{Small Recall} \\
\midrule
DiceCE         & 0.796 & 0.843 & 0.780 & 9.04 & -- \\
DiceCE\,+\,CAT & 0.823 & 0.839 & 0.781 & 8.68 & +0.027 [$-$0.010, +0.081] \\
DiceCE\,+\,MIL & 0.871 & 0.749 & 0.781 & 9.24 & \textbf{+0.076 [+0.033, +0.148]} \\
CATMIL         & 0.873 & 0.757 & 0.783 & 7.98 & \textbf{+0.077 [+0.030, +0.157]} \\
\bottomrule
\end{tabular}%
}
\end{table}

\textbf{The two terms act on different properties of the prediction.} The
MIL term carries the detection gain: alone, it raises small-lesion recall
by as much as CATMIL, and it also accounts for the lower lesion-wise F1, so
the additional false-positive components (\cref{sec:fp_pp}) are a
by-product of the detection signal. The CAT term alone moves small-lesion
recall in the same direction, close to the level of the other losses that
balance or reweight overlap (Tversky, FocalTversky, and CC-DiceCE,
0.831--0.834), though without a measurable difference from DiceCE and without
lowering lesion-wise F1. Combined, the two terms give the highest recall
together with the best mean DSC and HD95 of the three variants. This is
the division of labor the objective was designed for: the MIL term ensures
that each lesion is found, and the CAT term supervises the overlap of
lesions of different sizes in proportion.

\textbf{On this test set, the contribution of the CAT term is not
measurable.} CATMIL and DiceCE\,+\,MIL do not differ on any metric
(small-lesion recall +0.002 [$-0.012$, +0.019]; HD95 $-1.26$~mm
[$-6.82$, +1.55]). The loss-weight analysis shows that the ratio of the
two weights shifts the balance between voxel-level accuracy and
lesion-wise precision (\cref{appendix:catmil-loss-weights}), but those
runs are single models. Whether the CAT term contributes measurably to
delineation, or on data with a different lesion-size distribution, remains
to be tested.

\subsection{The Cost: Small False-Positive Components}
\label{sec:fp_pp}

\textbf{The false positives are many, small, and near true lesions.}
Without filtering, CATMIL produces 20.3 false-positive components per case
on MSLesSeg, against 5.4--7.0 for the voxel-level losses and 11.2 for
CC-DiceCE (\cref{appendix:lesion-level-analyses}). Yet its false-positive
volume is no larger than that of any baseline, and 70--80\% of the
false-positive voxels lie within a median of 1--2~mm of a reference lesion.
The cost is thus a set of small fragments and lesion-like foci, not large
false alarms. This is consistent with the ablation: a term that rewards
any confident voxel inside a lesion also lowers the threshold for
responding to small lesion-like structures elsewhere.

\textbf{Because the components are small, a size filter removes most of
them after training.} Removing predicted components smaller than 5 voxels
(PP5) raises CATMIL's lesion-wise F1 from 0.757 to 0.842, against gains of
0.013--0.015 for the voxel-level losses, while small-lesion recall
decreases only from 0.873 to 0.855 and remains the highest
(\cref{tab:loss_comparison_small_lesion_pp5} in
\cref{appendix:lesion-level-analyses}). After PP5 the lesion-wise F1
difference is no longer supported against any baseline on the test set,
whereas the recall gain remains in all six patients
(\cref{tab:paired_differences}). On the larger cross-validation sample,
however, a small F1 cost after PP5 remains (\cref{sec:catmil_oof}).

\subsubsection{Detection at Matched Lesion Precision}
\label{sec:matched_precision}

A fixed filter still leaves CATMIL at a lower precision than the
baselines. The stricter question is whether CATMIL detects more small
lesions at the \emph{same} precision, or only operates at a more liberal
point. Each loss was evaluated over a grid of probability thresholds
$t \in \{0.1,\dots,0.9\}$ and minimum component sizes $s \in [1, 170]$
voxels (\cref{fig:operating_grid}). For each baseline, the CATMIL setting
with the highest small-lesion recall whose precision is at least that of
the baseline at its default output was selected on MSLesSeg and applied
unchanged to 3D-MR-MS (\cref{tab:matched_precision}).

\begin{table}[t]
\centering
\caption{Small-lesion recall and lesion-wise F1 of CATMIL at matched
lesion-wise precision. For each baseline, the CATMIL setting $(t, s)$ with
the highest small-lesion recall whose precision on MSLesSeg is at least
that of the baseline at its default output $(0.5, 1)$ was selected on
MSLesSeg and applied unchanged to 3D-MR-MS. Values are differences CATMIL
minus baseline with 95\% patient-level bootstrap CIs in brackets;
differences whose CI excludes zero are shown in \textbf{bold}. The MSLesSeg
values are optimistic because the setting is selected on the same data.}
\label{tab:matched_precision}
\resizebox{\textwidth}{!}{%
\begin{tabular}{l r@{\hspace{0.4em}}l r@{\hspace{0.4em}}l r@{\hspace{0.4em}}l r@{\hspace{0.4em}}l}
\toprule
& \multicolumn{2}{c}{\textbf{vs.\ DiceCE}} & \multicolumn{2}{c}{\textbf{vs.\ Tversky}} & \multicolumn{2}{c}{\textbf{vs.\ FocalTversky}} & \multicolumn{2}{c}{\textbf{vs.\ CC-DiceCE}} \\
\cmidrule(lr){2-3}\cmidrule(lr){4-5}\cmidrule(lr){6-7}\cmidrule(lr){8-9}
CATMIL $(t, s)$ & \multicolumn{2}{c}{(0.1, 15)} & \multicolumn{2}{c}{(0.8, 4)} & \multicolumn{2}{c}{(0.2, 10)} & \multicolumn{2}{c}{(0.1, 5)} \\
\midrule
\multicolumn{9}{l}{\textit{MSLesSeg (selection)}} \\
\quad $\Delta$ Small Recall & \textbf{+0.048} & [+0.013, +0.111] & +0.016 & [$-$0.013, +0.060] & +0.016 & [$-$0.012, +0.057] & \textbf{+0.047} & [+0.020, +0.084] \\
\quad $\Delta$ Lesion F1    & \textbf{+0.021} & [+0.005, +0.041] & +0.007 & [$-$0.017, +0.028] & +0.007 & [$-$0.008, +0.023] & +0.015 & [$-$0.004, +0.034] \\
\midrule
\multicolumn{9}{l}{\textit{3D-MR-MS (transfer)}} \\
\quad $\Delta$ Small Recall & +0.040 & [$-$0.002, +0.087] & $-$0.010 & [$-$0.049, +0.024] & +0.006 & [$-$0.036, +0.043] & \textbf{+0.039} & [+0.003, +0.080] \\
\quad $\Delta$ Precision    & $-$0.050 & [$-$0.099, +0.001] & +0.030 & [$-$0.040, +0.094] & +0.006 & [$-$0.050, +0.067] & \textbf{$-$0.075} & [$-$0.129, $-$0.026] \\
\quad $\Delta$ Lesion F1    & $-$0.008 & [$-$0.036, +0.023] & +0.008 & [$-$0.020, +0.037] & +0.007 & [$-$0.022, +0.040] & $-$0.031 & [$-$0.072, +0.006] \\
\bottomrule
\end{tabular}%
}
\end{table}

\textbf{Against DiceCE and CC-DiceCE, CATMIL is a better detector, not only
a more liberal one.} At the precision of DiceCE, CATMIL detects more small
lesions (+0.048) and has a higher lesion-wise F1 (+0.021), both in all six
patients; at the precision of CC-DiceCE, it still detects more small
lesions (+0.047). The lower F1 of CATMIL at its default output is
therefore a matter of operating point. The transfer to 3D-MR-MS is weaker:
the recall gains keep their size (+0.040 and +0.039), but the precision is
not matched exactly on the new dataset, so part of the gain there again
comes from a more sensitive operating point.

\textbf{Against the Tversky-based losses, the advantage is largely one of
operating point.} At their precision, the recall gain shrinks to +0.016 on
MSLesSeg and to about zero on 3D-MR-MS, with CIs that include zero; tuned
each to its own maximum lesion-wise F1, the losses do not differ either.
Tversky and FocalTversky can thus reach much of CATMIL's sensitivity by
lowering their threshold or filter size. The advantage of CATMIL over them is
that it reaches the sensitive operating point by default, with the option
to trade sensitivity for precision after training.

\subsection{Robustness Beyond the Six Test Patients}
\label{sec:robustness}

The MSLesSeg test set has six patients. Three further analyses test
whether the findings depend on them: a larger sample from the same
dataset, a clinically defined lesion size, and a second dataset.

\subsubsection{Cross-Validation Patients}
\label{sec:catmil_oof}

The cross-validation predictions of the 81 scans of the 47 MSLesSeg
training patients were evaluated for all nine models, each scan predicted
by the fold model not trained on it. These predictions were not used for
any choice of the objective, whose weights were fixed beforehand.
Differences were assessed with a patient-level bootstrap over the 47
patients (\cref{tab:oof_paired}).

\begin{table}[t]
\centering
\caption{Paired differences between CATMIL and each baseline on the
cross-validation predictions of the 81 MSLesSeg training scans
(47 patients). Upper block: nnU-Net trained with the other losses; lower
block: the other architectures trained with DiceCE
(\cref{sec:arch_vs_objective}). Values are CATMIL minus baseline with the
95\% patient-level bootstrap CI (2{,}000 resamples) in brackets;
differences whose CI excludes zero are shown in \textbf{bold}. Lesion F1
is reported without filtering and after removing components smaller than
5 voxels (PP5).}
\label{tab:oof_paired}
\resizebox{\textwidth}{!}{%
\begin{tabular}{lccccc}
\toprule
\textbf{CATMIL minus} & \textbf{Small Lesion Recall} & \textbf{FN Count} & \textbf{Lesion F1} & \textbf{Lesion F1 (PP5)} & \textbf{DSC} \\
\midrule
DiceCE       & \textbf{+0.110 [+0.074, +0.140]} & \textbf{$-$3.37 [$-$4.58, $-$2.35]} & \textbf{$-$0.101 [$-$0.135, $-$0.067]} & \textbf{$-$0.025 [$-$0.044, $-$0.007]} & +0.004 [$-$0.002, +0.010] \\
Tversky      & \textbf{+0.070 [+0.037, +0.099]} & \textbf{$-$2.06 [$-$2.90, $-$1.34]} & \textbf{$-$0.100 [$-$0.137, $-$0.065]} & \textbf{$-$0.023 [$-$0.041, $-$0.005]} & $-$0.001 [$-$0.007, +0.005] \\
FocalTversky & \textbf{+0.073 [+0.039, +0.100]} & \textbf{$-$2.23 [$-$3.13, $-$1.45]} & \textbf{$-$0.097 [$-$0.128, $-$0.064]} & \textbf{$-$0.026 [$-$0.043, $-$0.009]} & $-$0.003 [$-$0.008, +0.003] \\
CC-DiceCE    & \textbf{+0.054 [+0.026, +0.077]} & \textbf{$-$1.65 [$-$2.32, $-$1.11]} & \textbf{$-$0.078 [$-$0.106, $-$0.050]} & \textbf{$-$0.026 [$-$0.043, $-$0.009]} & $-$0.002 [$-$0.009, +0.004] \\
\midrule
SegResNet (DiceCE) & \textbf{+0.107 [+0.076, +0.136]} & \textbf{$-$3.12 [$-$4.46, $-$2.10]} & \textbf{$-$0.098 [$-$0.131, $-$0.064]} & $-$0.018 [$-$0.037, +0.002] & +0.000 [$-$0.008, +0.009] \\
UMambaBot (DiceCE) & \textbf{+0.112 [+0.075, +0.144]} & \textbf{$-$3.53 [$-$4.96, $-$2.41]} & \textbf{$-$0.106 [$-$0.140, $-$0.072]} & \textbf{$-$0.030 [$-$0.051, $-$0.010]} & +0.001 [$-$0.008, +0.009] \\
SwinUNETR (DiceCE) & \textbf{+0.100 [+0.065, +0.132]} & \textbf{$-$3.28 [$-$4.80, $-$2.12]} & \textbf{$-$0.058 [$-$0.099, $-$0.015]} & +0.007 [$-$0.018, +0.032] & \textbf{+0.028 [+0.014, +0.044]} \\
UNETR (DiceCE)     & \textbf{+0.146 [+0.105, +0.187]} & \textbf{$-$4.59 [$-$6.49, $-$3.06]} & $-$0.009 [$-$0.055, +0.040] & \textbf{+0.057 [+0.026, +0.091]} & \textbf{+0.051 [+0.032, +0.074]} \\
\bottomrule
\end{tabular}%
}
\end{table}

\textbf{All three findings are reproduced, with narrower intervals.}
CATMIL raises small-lesion recall over every loss and every architecture,
from 0.692 with DiceCE to 0.802; CC-DiceCE again lies between the
voxel-level losses and CATMIL (0.747). The four architectures trained with
DiceCE again detect no more small lesions than nnU-Net, and the DSC of
CATMIL again does not differ from that of any other loss, although it
misses 1.65 to 3.37 fewer lesions per scan.

\textbf{The precision cost is also reproduced, and is estimated more
sharply.} After PP5, CATMIL's lesion-wise F1 remains slightly below that of
every other loss, including CC-DiceCE (by 0.023--0.026). The parity with
CC-DiceCE after PP5 on the test set is therefore not confirmed: on these
patients, CATMIL pays a small residual precision cost against all losses
while detecting more small lesions than all of them.

\subsubsection{Lesions of at Least 3~mm}
\label{sec:clinical_size}

The 150-voxel small-lesion threshold is a technical definition. Clinical
MRI reading counts lesions of at least 3~mm in longest
diameter~\cite{barkhof_2024_2025,filippi_mri_2016}, and the 2024 McDonald
criteria assess the central vein sign only in lesions of this
size~\cite{montalban_diagnosis_2025,barkhof_2024_2025}.
\Cref{tab:clinical_size} reports recall above and below this size.

\begin{table}[t]
\centering
\caption{Lesion recall by lesion size relative to the 3~mm longest
diameter used in clinical MRI reading. $n$: number of reference lesions.
Recall is the fraction of lesions detected, averaged over the five fold
models. $\Delta$: CATMIL minus DiceCE or CC-DiceCE with 95\% patient-level
bootstrap CI; differences whose CI excludes zero are shown in
\textbf{bold}.}
\label{tab:clinical_size}
\resizebox{\textwidth}{!}{%
\begin{tabular}{llrcccccll}
\toprule
\textbf{Dataset} & \textbf{Lesions} & $n$ & \textbf{CATMIL} & \textbf{DiceCE} & \textbf{Tversky} & \textbf{FocalTversky} & \textbf{CC-DiceCE} & $\Delta$ vs.\ DiceCE & $\Delta$ vs.\ CC-DiceCE \\
\midrule
MSLesSeg & $\geq$3~mm & 345 & 0.944 & 0.870 & 0.907 & 0.907 & 0.910 & \textbf{+0.074 [+0.032, +0.129]} & \textbf{+0.035 [+0.017, +0.057]} \\
         & $\geq$3~mm and $\leq$150 voxels & 233 & 0.924 & 0.827 & 0.874 & 0.875 & 0.884 & \textbf{+0.098 [+0.040, +0.171]} & \textbf{+0.040 [+0.020, +0.066]} \\
         & $<$3~mm & 16 & 0.275 & 0.087 & 0.175 & 0.225 & 0.213 & \textbf{+0.188 [+0.044, +0.300]} & +0.063 [$-$0.100, +0.188] \\
\midrule
3D-MR-MS & $\geq$3~mm & 269 & 0.894 & 0.830 & 0.837 & 0.833 & 0.863 & \textbf{+0.065 [+0.015, +0.125]} & \textbf{+0.031 [+0.010, +0.096]} \\
         & $\geq$3~mm and $\leq$150 voxels & 156 & 0.855 & 0.753 & 0.768 & 0.756 & 0.817 & \textbf{+0.103 [+0.050, +0.203]} & \textbf{+0.038 [+0.010, +0.129]} \\
         & $<$3~mm & 120 & 0.393 & 0.282 & 0.300 & 0.308 & 0.350 & \textbf{+0.112 [+0.010, +0.167]} & +0.043 [$-$0.013, +0.075] \\
\bottomrule
\end{tabular}%
}
\end{table}

\textbf{The gain extends to the lesions a reader would count.} For lesions
of at least 3~mm, CATMIL raises recall over DiceCE on both datasets
(0.870 to 0.944 on MSLesSeg, 0.830 to 0.894 on 3D-MR-MS) and over
CC-DiceCE by about 3 percentage points, all with CIs excluding zero. The
largest gain, about 10 percentage points over DiceCE on both datasets, is
in lesions that are small but above the clinical size, which are the
lesions most relevant for counting and for central vein sign assessment
and those that a voxel-wise objective tends to miss. Below 3~mm, recall
stays low for every loss (at most 0.39). Such lesions can still be
considered demyelinating in the corpus callosum, the floor of the fourth
ventricle, and the outer margin of the pons~\cite{barkhof_2024_2025}, but
lesion location was not analyzed here.

\subsubsection{A Second Dataset: 3D-MR-MS}
\label{sec:3dmrms}

3D-MR-MS was acquired at a different center with a finer, anisotropic
resolution and a much larger share of very small lesions, and all five
losses were trained on it with the settings fixed on MSLesSeg.

\textbf{The detection gain transfers.} CATMIL again detects the most small
lesions (0.618 against 0.537 for DiceCE) and misses 23\% fewer lesions,
in five of the six patients against every baseline
(\cref{tab:loss_comparison_small_lesion,tab:paired_differences}).
CC-DiceCE is again the closest competitor: CATMIL misses fewer lesions
than it (2.27 per case, CI excluding zero), although the small-lesion
recall difference (+0.039) narrowly includes zero. The lesion-level
analyses reproduce the mechanism: the exchange of detected lesions with
DiceCE is 33 to 2.4 per fold model, and the fraction of lesions of at most
10~mm$^3$ with no response falls from 57--58\% to 39\%
(\cref{appendix:lesion-level-analyses}).

\textbf{The cost is larger, and the five-voxel filter does not suffice.}
Without filtering, CATMIL produces about 150 false-positive components per
case, against 14--19 for the voxel-level losses, and its lesion-wise F1 is
the lowest in every patient. Its DSC is also slightly lower than that of
DiceCE ($-0.012$) in all six patients, so on this dataset the DSC is
slightly reduced rather than preserved. After PP5, the F1 gap narrows but
remains against the voxel-level losses, and the recall advantage survives
only against DiceCE. The false-positive components on this dataset are
small in voxels rather than in millimeters, and a 20-voxel filter brings
CATMIL to the false-positive count of DiceCE with slightly higher recall,
but this setting was chosen after seeing the data. One test patient
contributes 249 of the 389 reference lesions, so pooled lesion counts on
this dataset are dominated by that patient, whereas the patient-level
analysis weights patients equally.

\section{Discussion}
\label{sec:discussion}
\paragraph{Why standard objectives miss small lesions.}
The results point to a specific failure. The baselines do not miss small
lesions narrowly; for most of them the network produces no response at all,
so the omission is fixed during training and cannot be undone at
inference. Four other architectures trained with the same loss show the
same omission, while a change of objective with the network held fixed
removes much of it. Within the range of networks tested, the omission is
therefore better explained by the supervision than by the capacity of the
network. \Cref{eq:sl_voxel_loss} gives the reason: a lesion of a few voxels
contributes a few terms to a sum over hundreds of thousands, and an
optimizer has little incentive to fit it.

\paragraph{Why both balance and detection.}
The two terms of CATMIL answer two distinct deficits of a voxel-wise loss.
Balance alone, whether through the CAT term, CC-DiceCE, or a heavier
penalty on false-negative voxels, raises small-lesion recall only part of
the way (\cref{sec:ablation}). The form of the terms suggests why. In an
overlap term, even one balanced across lesions, a lesion enters through its
soft recall $r_k$ linearly (\cref{eq:sl_cat_lesion_form}): a completely
missed lesion and a half-segmented lesion of the same weight receive the
same marginal gradient, so nothing in the objective marks a missed lesion
as a different kind of error. The MIL term does. Its contribution,
$-\log s_k$, grows steeply as the strongest response in a lesion approaches
zero, its gradient $1/(K s_k)$ is largest for exactly those lesions, and it
vanishes once a single voxel is confident. Conversely, the MIL term says
nothing about how much of a lesion is segmented; that is left to the
overlap terms, and the CAT term ensures that this delineation signal is
not dominated by large lesions. Detection and delineation are thereby
supervised separately. This division is visible in the results: the
recovered lesions are detected but often only partly delineated, and the
gain persists, though narrowing, under stricter coverage criteria
(\cref{appendix:strict-detection}). In the present ablation the detection
gain is carried by the MIL term, and a measurable benefit of the CAT term
on delineation was not shown.

This interpretation has a limit. Against the Tversky-based losses, much of
the gain disappears at matched precision (\cref{sec:matched_precision}), so
those losses can reach a similar sensitivity by a lower threshold or filter
size. What the lesion-level terms change reliably is where the default operating point lies, and,
against DiceCE and CC-DiceCE, the detection achievable at a given
precision.

\paragraph{Relation to instance-level losses.}
Instance-level losses~\cite{shirokikh_universal_2020,kofler_blob_2023,bouteille_learning_2026}
were developed for the same problem and do improve small-lesion recall;
CC-DiceCE did so here, at some cost in precision, as its authors
reported~\cite{bouteille_learning_2026}. CATMIL is complementary rather
than an alternative to them: its CAT term follows the same balancing
principle, and the MIL term adds what balancing lacks, a lesion-level
existence signal that is independent of overlap. Combining this signal
with instance-level losses other than the CAT term, or with
BiCC~\cite{bouteille_bicc_2026}, which penalizes unmatched predicted
components, is a direct next step.

\paragraph{The trade-off and its control.}
The detection gain and the false-positive cost arise from the same term:
a loss that rewards one confident voxel inside every lesion also makes the
network respond to small lesion-like foci elsewhere. The resulting
components are small and mostly close to true lesions, so they add little
false-positive volume and can be removed by a size filter after training.
This makes the operating point adjustable without retraining. The filter
size, however, depends on the data: a five-voxel filter sufficed on
MSLesSeg but not on 3D-MR-MS, and a small residual precision cost remains
on the MSLesSeg cross-validation patients. Moving this control into the
objective, for instance by penalizing predicted components that overlap no
reference lesion, would act on exactly the error that the MIL term
introduces.

\paragraph{Clinical implications.}
In MS, a lesion that is never detected cannot be counted, whereas a
false-positive component can be dismissed when a reader reviews the
segmentation. Where automatic segmentation supports rather than replaces
the radiologist, trading missed lesions for small false positives is
therefore a favorable direction, with the filter size chosen for the use:
small for screening and follow-up, larger where a clean lesion map is
needed. The gain is not confined to lesions below the size used in
clinical reading. It is largest for small lesions of at least 3~mm, which
are the lesions that are counted and in which the central vein sign, a
diagnostic marker in the 2024 McDonald criteria, can be
assessed~\cite{montalban_diagnosis_2025,barkhof_2024_2025}.

\paragraph{Limitations.}
First, the evaluation is restricted to MS, and the objective has not been
tested on other pathologies. Second, CATMIL was trained only with nnU-Net;
the other architectures were trained with DiceCE, with default
hyperparameters, a 150-epoch budget, and only on MSLesSeg, so it is not
known whether CATMIL raises their recall as much. Among instance-level
losses, only CC-DiceCE was trained under the same protocol, using the
authors' loss within this pipeline rather than their own setup. Third, the
test sets are small (six patients each), so no paired test can reach
$p < 0.05$ after correction, and the analysis rests on patient-level CIs
and on consistency across patients; on 3D-MR-MS, one patient also
contributes 64\% of the reference lesions. The 47 cross-validation
patients of MSLesSeg reproduce the main findings, but the corresponding
analysis was not performed on 3D-MR-MS. Fourth, the operating points at
matched precision were selected on the MSLesSeg test set, so only their
transfer to 3D-MR-MS is free of selection. Fifth, the ablation was run on
MSLesSeg only, where the contribution of the CAT term was not measurable;
its role should be tested on further datasets.
Sixth, connected components are computed within each training patch, so a
lesion cut by the patch border is weighted by its visible part. Finally,
lesions of a few voxels and lesions with low contrast remain difficult for
every loss: the objective changes how much each lesion counts during
training, but it cannot create evidence that the image does not contain.

\section{Conclusion}
\label{sec:conclusion}
This work asked why segmentation models miss small lesions in brain MRI.
In MS lesion segmentation, standard objectives produce almost no response
to most of the small lesions they miss, and the architectures we tested do
not change this when trained with the same objective. The omission arises
in training, from how supervision is distributed across lesions.

CATMIL supervises each lesion as a unit in two ways: the CAT term balances
the overlap signal across lesions of different sizes, and the MIL term
penalizes every lesion in which no voxel is detected. Together they detect
more small lesions than losses that only reweight voxels or balance
lesions within an overlap term, with the detection gain carried mainly by
the MIL term. The gain holds
across patients, on a larger cross-validation sample, for lesions of
clinically reportable size, and on a second dataset trained with the same
loss weights, and it does not require any change to the network.

The cost is additional small false-positive components. They can be
removed after training, and at the precision of the standard loss CATMIL
still detects more small lesions, but the filter size depends on the data,
and against Tversky-based losses part of the gain reflects a more
sensitive operating point. CATMIL is therefore most useful where a missed
lesion is a more serious error than a small false positive. Separating
detection from delineation in the objective suggests two next steps: an
analogous term that penalizes unmatched predicted components, and
lesion-level supervision of complete delineation rather than of a single
detected voxel.

\paragraph{Acknowledgements.}
This work was supported by a grant from the Russian Science Foundation (RSF 23-15-00377-$\pi$). The authors used ChatGPT (OpenAI) to improve the language and readability of this manuscript. All content was reviewed and approved by the authors.

\bibliographystyle{unsrtnat}
\bibliography{references}

@article{lesjak_novel_2018,
	title = {A {Novel} {Public} {MR} {Image} {Dataset} of {Multiple} {Sclerosis} {Patients} {With} {Lesion} {Segmentations} {Based} on {Multi}-rater {Consensus}},
	volume = {16},
	issn = {1539-2791, 1559-0089},
	url = {http://link.springer.com/10.1007/s12021-017-9348-7},
	doi = {10.1007/s12021-017-9348-7},
	language = {en},
	number = {1},
	urldate = {2026-05-24},
	journal = {Neuroinformatics},
	author = {Lesjak, Žiga and Galimzianova, Alfiia and Koren, Aleš and Lukin, Matej and Pernuš, Franjo and Likar, Boštjan and Špiclin, Žiga},
	month = jan,
	year = {2018},
	pages = {51--63},
}

@misc{hatamizadeh_swin_2022,
	title = {Swin {UNETR}: {Swin} {Transformers} for {Semantic} {Segmentation} of {Brain} {Tumors} in {MRI} {Images}},
	copyright = {Creative Commons Attribution Non Commercial No Derivatives 4.0 International},
	shorttitle = {Swin {UNETR}},
	url = {https://arxiv.org/abs/2201.01266},
	doi = {10.48550/ARXIV.2201.01266},
	urldate = {2026-04-09},
	publisher = {arXiv},
	author = {Hatamizadeh, Ali and Nath, Vishwesh and Tang, Yucheng and Yang, Dong and Roth, Holger and Xu, Daguang},
	year = {2022},
	note = {Version Number: 1},
}

@misc{hatamizadeh_unetr_2021,
	title = {{UNETR}: {Transformers} for {3D} {Medical} {Image} {Segmentation}},
	copyright = {Creative Commons Attribution Non Commercial No Derivatives 4.0 International},
	shorttitle = {{UNETR}},
	url = {https://arxiv.org/abs/2103.10504},
	doi = {10.48550/ARXIV.2103.10504},
	urldate = {2026-04-09},
	publisher = {arXiv},
	author = {Hatamizadeh, Ali and Tang, Yucheng and Nath, Vishwesh and Yang, Dong and Myronenko, Andriy and Landman, Bennett and Roth, Holger and Xu, Daguang},
	year = {2021},
	note = {Version Number: 3},
}

@article{isensee_nnu-net_2021,
	title = {{nnU}-{Net}: a self-configuring method for deep learning-based biomedical image segmentation},
	volume = {18},
	issn = {1548-7091, 1548-7105},
	shorttitle = {{nnU}-{Net}},
	url = {https://www.nature.com/articles/s41592-020-01008-z},
	doi = {10.1038/s41592-020-01008-z},
	language = {en},
	number = {2},
	urldate = {2026-04-09},
	journal = {Nature Methods},
	author = {Isensee, Fabian and Jaeger, Paul F. and Kohl, Simon A. A. and Petersen, Jens and Maier-Hein, Klaus H.},
	month = feb,
	year = {2021},
	pages = {203--211},
}

@incollection{navab_u-net_2015,
	address = {Cham},
	title = {U-{Net}: {Convolutional} {Networks} for {Biomedical} {Image} {Segmentation}},
	volume = {9351},
	isbn = {978-3-319-24573-7 978-3-319-24574-4},
	shorttitle = {U-{Net}},
	url = {http://link.springer.com/10.1007/978-3-319-24574-4_28},
	doi = {10.1007/978-3-319-24574-4_28},
	language = {en},
	urldate = {2026-04-09},
	booktitle = {Medical {Image} {Computing} and {Computer}-{Assisted} {Intervention} – {MICCAI} 2015},
	publisher = {Springer International Publishing},
	author = {Ronneberger, Olaf and Fischer, Philipp and Brox, Thomas},
	editor = {Navab, Nassir and Hornegger, Joachim and Wells, William M. and Frangi, Alejandro F.},
	year = {2015},
	note = {Series Title: Lecture Notes in Computer Science},
	pages = {234--241},
}

@article{pal_deep_2021,
	title = {Deep multiple-instance learning for abnormal cell detection in cervical histopathology images},
	volume = {138},
	issn = {00104825},
	url = {https://linkinghub.elsevier.com/retrieve/pii/S0010482521006843},
	doi = {10.1016/j.compbiomed.2021.104890},
	language = {en},
	urldate = {2026-04-05},
	journal = {Computers in Biology and Medicine},
	author = {Pal, Anabik and Xue, Zhiyun and Desai, Kanan and Aina F Banjo, Adekunbiola and Adepiti, Clement Akinfolarin and Long, L. Rodney and Schiffman, Mark and Antani, Sameer},
	month = nov,
	year = {2021},
	pages = {104890},
}

@article{huang_geometric_2026,
	title = {Geometric multi-instance learning for weakly supervised gastric cancer segmentation},
	volume = {9},
	issn = {2398-6352},
	url = {https://www.nature.com/articles/s41746-025-02287-6},
	doi = {10.1038/s41746-025-02287-6},
	language = {en},
	number = {1},
	urldate = {2026-04-05},
	journal = {npj Digital Medicine},
	author = {Huang, Chenshen and Xia, Haoyun and Xiao, Xi and Chen, Hong and Jiang, Yiqing and Lyu, Yahui and Ni, Zhizhan and Wang, Tianyang and Wang, Ning and Huang, Qi},
	month = jan,
	year = {2026},
	pages = {101},
}

@article{liu_object_2026,
	title = {Object knowledge-aware multiple instance learning for small tumor segmentation},
	volume = {115},
	issn = {17468094},
	url = {https://linkinghub.elsevier.com/retrieve/pii/S1746809425019111},
	doi = {10.1016/j.bspc.2025.109400},
	language = {en},
	urldate = {2026-04-05},
	journal = {Biomedical Signal Processing and Control},
	author = {Liu, Haofeng and Gou, Shuiping and Zhou, Yanyan and Jiao, Changzhe and Liu, Wenbo and Shi, Mei and Luo, Zhonghua},
	month = apr,
	year = {2026},
	pages = {109400},
}

@article{aqib_javed_novel_2025,
	title = {A novel regularization approach for loss functions to reduce instance imbalance in biomedical image segmentation},
	volume = {119},
	issn = {14769271},
	url = {https://linkinghub.elsevier.com/retrieve/pii/S1476927125002154},
	doi = {10.1016/j.compbiolchem.2025.108555},
	language = {en},
	urldate = {2026-04-05},
	journal = {Computational Biology and Chemistry},
	author = {Aqib Javed, Muhammad and Khuram Shahzad, Muhammad and Syed Muhammad Bilal Ali, Hafiz},
	month = dec,
	year = {2025},
	pages = {108555},
}

@article{zhang_all-net_2021,
	title = {{ALL}-{Net}: {Anatomical} information lesion-wise loss function integrated into neural network for multiple sclerosis lesion segmentation},
	volume = {32},
	issn = {22131582},
	shorttitle = {{ALL}-{Net}},
	url = {https://linkinghub.elsevier.com/retrieve/pii/S2213158221002989},
	doi = {10.1016/j.nicl.2021.102854},
	language = {en},
	urldate = {2026-04-05},
	journal = {NeuroImage: Clinical},
	author = {Zhang, Hang and Zhang, Jinwei and Li, Chao and Sweeney, Elizabeth M. and Spincemaille, Pascal and Nguyen, Thanh D. and Gauthier, Susan A. and Wang, Yi and Marcille, Melanie},
	year = {2021},
	pages = {102854},
}

@misc{azad_loss_2023,
	title = {Loss {Functions} in the {Era} of {Semantic} {Segmentation}: {A} {Survey} and {Outlook}},
	copyright = {Creative Commons Attribution 4.0 International},
	shorttitle = {Loss {Functions} in the {Era} of {Semantic} {Segmentation}},
	url = {https://arxiv.org/abs/2312.05391},
	doi = {10.48550/ARXIV.2312.05391},
	urldate = {2026-04-05},
	publisher = {arXiv},
	author = {Azad, Reza and Heidary, Moein and Yilmaz, Kadir and Hüttemann, Michael and Karimijafarbigloo, Sanaz and Wu, Yuli and Schmeink, Anke and Merhof, Dorit},
	year = {2023},
	note = {Version Number: 1},
}

@article{valverde_region-wise_2023,
	title = {Region-wise loss for biomedical image segmentation},
	volume = {136},
	issn = {00313203},
	url = {https://linkinghub.elsevier.com/retrieve/pii/S0031320322006872},
	doi = {10.1016/j.patcog.2022.109208},
	language = {en},
	urldate = {2026-04-04},
	journal = {Pattern Recognition},
	author = {Valverde, Juan Miguel and Tohka, Jussi},
	month = apr,
	year = {2023},
	pages = {109208},
}

@article{chen_adaptive_2023,
	title = {Adaptive {Region}-{Specific} {Loss} for {Improved} {Medical} {Image} {Segmentation}},
	volume = {45},
	issn = {1939-3539},
	doi = {10.1109/TPAMI.2023.3289667},
	language = {eng},
	number = {11},
	journal = {IEEE transactions on pattern analysis and machine intelligence},
	author = {Chen, Yizheng and Yu, Lequan and Wang, Jen-Yeu and Panjwani, Neil and Obeid, Jean-Pierre and Liu, Wu and Liu, Lianli and Kovalchuk, Nataliya and Gensheimer, Michael Francis and Vitzthum, Lucas Kas and Beadle, Beth M. and Chang, Daniel T. and Le, Quynh-Thu and Han, Bin and Xing, Lei},
	month = nov,
	year = {2023},
	pages = {13408--13421},
}

@article{hashemi_asymmetric_2019,
	title = {Asymmetric {Loss} {Functions} and {Deep} {Densely}-{Connected} {Networks} for {Highly}-{Imbalanced} {Medical} {Image} {Segmentation}: {Application} to {Multiple} {Sclerosis} {Lesion} {Detection}},
	volume = {7},
	copyright = {https://ieeexplore.ieee.org/Xplorehelp/downloads/license-information/OAPA.html},
	issn = {2169-3536},
	shorttitle = {Asymmetric {Loss} {Functions} and {Deep} {Densely}-{Connected} {Networks} for {Highly}-{Imbalanced} {Medical} {Image} {Segmentation}},
	url = {https://ieeexplore.ieee.org/document/8573779/},
	doi = {10.1109/ACCESS.2018.2886371},
	urldate = {2026-04-04},
	journal = {IEEE Access},
	author = {Hashemi, Seyed Raein and Mohseni Salehi, Seyed Sadegh and Erdogmus, Deniz and Prabhu, Sanjay P. and Warfield, Simon K. and Gholipour, Ali},
	year = {2019},
	pages = {1721--1735},
}

@article{dereskewicz_flames_2025,
	title = {{FLAMeS}: {A} {Robust} {Deep} {Learning} {Model} for {Automated} {Multiple} {Sclerosis} {Lesion} {Segmentation}},
	shorttitle = {{FLAMeS}},
	doi = {10.1101/2025.05.19.25327707},
	language = {eng},
	journal = {medRxiv: The Preprint Server for Health Sciences},
	author = {Dereskewicz, Emma and La Rosa, Francesco and Dos Santos Silva, Jonadab and Sizer, Edward and Kohli, Amit and Wynen, Maxence and Mullins, William A. and Maggi, Pietro and Levy, Sarah and Onyemeh, Kamso and Ayci, Batuhan and Solomon, Andrew J. and Assländer, Jakob and Al-Louzi, Omar and Reich, Daniel S. and Sumowski, James and Beck, Erin S.},
	month = may,
	year = {2025},
	pages = {2025.05.19.25327707},
}

@article{yeung_calibrating_2023,
	title = {Calibrating the {Dice} {Loss} to {Handle} {Neural} {Network} {Overconfidence} for {Biomedical} {Image} {Segmentation}},
	volume = {36},
	issn = {1618-727X},
	doi = {10.1007/s10278-022-00735-3},
	language = {eng},
	number = {2},
	journal = {Journal of Digital Imaging},
	author = {Yeung, Michael and Rundo, Leonardo and Nan, Yang and Sala, Evis and Schönlieb, Carola-Bibiane and Yang, Guang},
	month = apr,
	year = {2023},
	pages = {739--752},
}

@misc{lin_focal_2017,
	title = {Focal {Loss} for {Dense} {Object} {Detection}},
	copyright = {arXiv.org perpetual, non-exclusive license},
	url = {https://arxiv.org/abs/1708.02002},
	doi = {10.48550/ARXIV.1708.02002},
	urldate = {2026-04-04},
	publisher = {arXiv},
	author = {Lin, Tsung-Yi and Goyal, Priya and Girshick, Ross and He, Kaiming and Dollár, Piotr},
	year = {2017},
	note = {Version Number: 2},
}

@inproceedings{abraham_novel_2019,
	address = {Venice, Italy},
	title = {A {Novel} {Focal} {Tversky} {Loss} {Function} {With} {Improved} {Attention} {U}-{Net} for {Lesion} {Segmentation}},
	copyright = {https://ieeexplore.ieee.org/Xplorehelp/downloads/license-information/Crown.html},
	isbn = {978-1-5386-3641-1},
	url = {https://ieeexplore.ieee.org/document/8759329/},
	doi = {10.1109/ISBI.2019.8759329},
	urldate = {2026-04-04},
	booktitle = {2019 {IEEE} 16th {International} {Symposium} on {Biomedical} {Imaging} ({ISBI} 2019)},
	publisher = {IEEE},
	author = {Abraham, Nabila and Khan, Naimul Mefraz},
	month = apr,
	year = {2019},
	pages = {683--687},
}

@article{wiltgen_lst-ai_2024,
	title = {{LST}-{AI}: {A} deep learning ensemble for accurate {MS} lesion segmentation},
	volume = {42},
	issn = {22131582},
	shorttitle = {{LST}-{AI}},
	url = {https://linkinghub.elsevier.com/retrieve/pii/S2213158224000500},
	doi = {10.1016/j.nicl.2024.103611},
	language = {en},
	urldate = {2026-04-04},
	journal = {NeuroImage: Clinical},
	author = {Wiltgen, Tun and McGinnis, Julian and Schlaeger, Sarah and Kofler, Florian and Voon, CuiCi and Berthele, Achim and Bischl, Daria and Grundl, Lioba and Will, Nikolaus and Metz, Marie and Schinz, David and Sepp, Dominik and Prucker, Philipp and Schmitz-Koep, Benita and Zimmer, Claus and Menze, Bjoern and Rueckert, Daniel and Hemmer, Bernhard and Kirschke, Jan and Mühlau, Mark and Wiestler, Benedikt},
	year = {2024},
	pages = {103611},
}

@article{wang_survey_1997,
	title = {Survey of the distribution of lesion size in multiple sclerosis: implication for the measurement of total lesion load},
	volume = {63},
	issn = {0022-3050},
	shorttitle = {Survey of the distribution of lesion size in multiple sclerosis},
	doi = {10.1136/jnnp.63.4.452},
	language = {eng},
	number = {4},
	journal = {Journal of Neurology, Neurosurgery, and Psychiatry},
	author = {Wang, L. and Lai, H. M. and Thompson, A. J. and Miller, D. H.},
	month = oct,
	year = {1997},
	pages = {452--455},
}

@article{thompson_diagnosis_2018,
	title = {Diagnosis of multiple sclerosis: 2017 revisions of the {McDonald} criteria},
	volume = {17},
	issn = {14744422},
	shorttitle = {Diagnosis of multiple sclerosis},
	url = {https://linkinghub.elsevier.com/retrieve/pii/S1474442217304702},
	doi = {10.1016/S1474-4422(17)30470-2},
	language = {en},
	number = {2},
	urldate = {2026-03-28},
	journal = {The Lancet Neurology},
	author = {Thompson, Alan J and Banwell, Brenda L and Barkhof, Frederik and Carroll, William M and Coetzee, Timothy and Comi, Giancarlo and Correale, Jorge and Fazekas, Franz and Filippi, Massimo and Freedman, Mark S and Fujihara, Kazuo and Galetta, Steven L and Hartung, Hans Peter and Kappos, Ludwig and Lublin, Fred D and Marrie, Ruth Ann and Miller, Aaron E and Miller, David H and Montalban, Xavier and Mowry, Ellen M and Sorensen, Per Soelberg and Tintoré, Mar and Traboulsee, Anthony L and Trojano, Maria and Uitdehaag, Bernard M J and Vukusic, Sandra and Waubant, Emmanuelle and Weinshenker, Brian G and Reingold, Stephen C and Cohen, Jeffrey A},
	month = feb,
	year = {2018},
	pages = {162--173},
}

@article{taha_metrics_2015,
	title = {Metrics for evaluating {3D} medical image segmentation: analysis, selection, and tool},
	volume = {15},
	issn = {1471-2342},
	shorttitle = {Metrics for evaluating {3D} medical image segmentation},
	url = {http://bmcmedimaging.biomedcentral.com/articles/10.1186/s12880-015-0068-x},
	doi = {10.1186/s12880-015-0068-x},
	language = {en},
	number = {1},
	urldate = {2026-03-15},
	journal = {BMC Medical Imaging},
	author = {Taha, Abdel Aziz and Hanbury, Allan},
	month = dec,
	year = {2015},
	pages = {29},
}

@article{guarnera_mslesseg_2025,
	title = {{MSLesSeg}: baseline and benchmarking of a new {Multiple} {Sclerosis} {Lesion} {Segmentation} dataset},
	volume = {12},
	issn = {2052-4463},
	shorttitle = {{MSLesSeg}},
	url = {https://www.nature.com/articles/s41597-025-05250-y},
	doi = {10.1038/s41597-025-05250-y},
	language = {en},
	number = {1},
	urldate = {2026-03-05},
	journal = {Scientific Data},
	author = {Guarnera, Francesco and Rondinella, Alessia and Crispino, Elena and Russo, Giulia and Di Lorenzo, Clara and Maimone, Davide and Pappalardo, Francesco and Battiato, Sebastiano},
	month = may,
	year = {2025},
	pages = {920},
}

@misc{ma_u-mamba_2024,
	title = {U-{Mamba}: {Enhancing} {Long}-range {Dependency} for {Biomedical} {Image} {Segmentation}},
	copyright = {Creative Commons Attribution Non Commercial No Derivatives 4.0 International},
	shorttitle = {U-{Mamba}},
	url = {https://arxiv.org/abs/2401.04722},
	doi = {10.48550/ARXIV.2401.04722},
	urldate = {2026-03-05},
	publisher = {arXiv},
	author = {Ma, Jun and Li, Feifei and Wang, Bo},
	year = {2024},
	note = {Version Number: 1},
}

@misc{salehi_tversky_2017,
	title = {Tversky loss function for image segmentation using {3D} fully convolutional deep networks},
	copyright = {arXiv.org perpetual, non-exclusive license},
	url = {https://arxiv.org/abs/1706.05721},
	doi = {10.48550/ARXIV.1706.05721},
	urldate = {2026-03-05},
	publisher = {arXiv},
	author = {Salehi, Seyed Sadegh Mohseni and Erdogmus, Deniz and Gholipour, Ali},
	year = {2017},
	note = {Version Number: 1},
}

@misc{bouteille_learning_2026,
	title = {Learning to {Look} {Closer}: {A} {New} {Instance}-{Wise} {Loss} for {Small} {Cerebral} {Lesion} {Segmentation}},
	url = {https://arxiv.org/abs/2511.17146},
	doi = {10.48550/arXiv.2511.17146},
	publisher = {arXiv},
	author = {Bouteille, Luc and Jaus, Alexander and Kleesiek, Jens and Stiefelhagen, Rainer and Heine, Lukas},
	year = {2026},
	note = {Accepted at IEEE ISBI 2026},
}

@article{dice_measures_1945,
	title = {Measures of the {Amount} of {Ecologic} {Association} {Between} {Species}},
	volume = {26},
	issn = {0012-9658},
	doi = {10.2307/1932409},
	number = {3},
	journal = {Ecology},
	author = {Dice, Lee R.},
	year = {1945},
	pages = {297--302},
}

@article{dietterich_solving_1997,
	title = {Solving the multiple instance problem with axis-parallel rectangles},
	volume = {89},
	issn = {00043702},
	doi = {10.1016/S0004-3702(96)00034-3},
	number = {1-2},
	journal = {Artificial Intelligence},
	author = {Dietterich, Thomas G. and Lathrop, Richard H. and Lozano-Pérez, Tomás},
	month = jan,
	year = {1997},
	pages = {31--71},
}

@inproceedings{hsu_weakly_2019,
	title = {Weakly {Supervised} {Instance} {Segmentation} using the {Bounding} {Box} {Tightness} {Prior}},
	url = {https://papers.nips.cc/paper_files/paper/2019/hash/e6e713296627dff6475085cc6a224464-Abstract.html},
	booktitle = {Advances in {Neural} {Information} {Processing} {Systems}},
	volume = {32},
	publisher = {Curran Associates, Inc.},
	author = {Hsu, Cheng-Chun and Hsu, Kuang-Jui and Tsai, Chung-Chi and Lin, Yen-Yu and Chuang, Yung-Yu},
	year = {2019},
}

@misc{ilse_attention-based_2018,
	title = {Attention-based {Deep} {Multiple} {Instance} {Learning}},
	url = {http://arxiv.org/abs/1802.04712},
	doi = {10.48550/arXiv.1802.04712},
	urldate = {2026-05-26},
	publisher = {arXiv},
	author = {Ilse, Maximilian and Tomczak, Jakub M. and Welling, Max},
	month = jun,
	year = {2018},
	note = {arXiv:1802.04712 [cs.LG]},
}

@article{jaus_every_2025,
	title = {Every {Component} {Counts}: {Rethinking} the {Measure} of {Success} for {Medical} {Semantic} {Segmentation} in {Multi}-{Instance} {Segmentation} {Tasks}},
	volume = {39},
	doi = {10.1609/aaai.v39i4.32408},
	number = {4},
	journal = {Proceedings of the AAAI Conference on Artificial Intelligence},
	author = {Jaus, Alexander and Seibold, Constantin Marc and Reiß, Simon and Marinov, Zdravko and Li, Keyi and Ye, Zeling and Krieg, Stefan and Kleesiek, Jens and Stiefelhagen, Rainer},
	month = apr,
	year = {2025},
	pages = {3904--3912},
}

@inproceedings{kofler_blob_2023,
	address = {Cham},
	title = {Blob Loss: {Instance} {Imbalance} {Aware} {Loss} {Functions} for {Semantic} {Segmentation}},
	isbn = {978-3-031-34048-2},
	url = {https://doi.org/10.1007/978-3-031-34048-2_58},
	doi = {10.1007/978-3-031-34048-2_58},
	urldate = {2026-09-12},
	booktitle = {Information {Processing} in {Medical} {Imaging}: 28th {International} {Conference}, {IPMI} 2023, {San} {Carlos} de {Bariloche}, {Argentina}, {June} 18–23, 2023, {Proceedings}},
	volume = {13939},
	series = {Lecture {Notes} in {Computer} {Science}},
	publisher = {Springer},
	author = {Kofler, Florian and Shit, Suprosanna and Ezhov, Ivan and Fidon, Lucas and Horvath, Izabela and Al-Maskari, Rami and Li, Hongwei Bran and Bhatia, Harsharan and Loehr, Timo and Piraud, Marie and Ertürk, Ali and Kirschke, Jan and Peeken, Jan C. and Vercauteren, Tom and Zimmer, Claus and Wiestler, Benedikt and Menze, Bjoern},
	year = {2023},
	pages = {755--767},
	note = {Preprint available as arXiv:2205.08209},
}

@article{ma_loss_2021,
	title = {Loss odyssey in medical image segmentation},
	volume = {71},
	issn = {13618415},
	doi = {10.1016/j.media.2021.102035},
	journal = {Medical Image Analysis},
	author = {Ma, Jun and Chen, Jianan and Ng, Matthew and Huang, Rui and Li, Yu and Li, Chen and Yang, Xiaoping and Martel, Anne L.},
	month = jul,
	year = {2021},
	pages = {102035},
}

@inproceedings{pinheiro_image-level_2015,
	title = {From image-level to pixel-level labeling with {Convolutional} {Networks}},
	doi = {10.1109/CVPR.2015.7298780},
	booktitle = {2015 {IEEE} {Conference} on {Computer} {Vision} and {Pattern} {Recognition} ({CVPR})},
	publisher = {IEEE},
	author = {Pinheiro, Pedro O. and Collobert, Ronan},
	month = jun,
	year = {2015},
	pages = {1713--1721},
}

@inproceedings{rachmadi_improving_2024,
	title = {Improving {Segmentation} of {Objects} with {Varying} {Sizes} in {Biomedical} {Images} using {Instance}-wise and {Center}-of-{Instance} {Segmentation} {Loss} {Function}},
	url = {https://proceedings.mlr.press/v227/rachmadi24a.html},
	booktitle = {Medical {Imaging} with {Deep} {Learning}},
	volume = {227},
	series = {Proceedings of {Machine} {Learning} {Research}},
	publisher = {PMLR},
	author = {Rachmadi, Muhammad Febrian and Poon, Charissa and Skibbe, Henrik},
	year = {2024},
	pages = {286--300},
}

@inproceedings{shirokikh_universal_2020,
	address = {Cham},
	title = {Universal {Loss} {Reweighting} to {Balance} {Lesion} {Size} {Inequality} in {3D} {Medical} {Image} {Segmentation}},
	isbn = {978-3-030-59719-1},
	doi = {10.1007/978-3-030-59719-1_51},
	booktitle = {Medical {Image} {Computing} and {Computer} {Assisted} {Intervention} – {MICCAI} 2020},
	volume = {12264},
	series = {Lecture {Notes} in {Computer} {Science}},
	publisher = {Springer},
	author = {Shirokikh, Boris and Shevtsov, Alexey and Kurmukov, Anvar and Dalechina, Alexandra and Krivov, Egor and Kostjuchenko, Valery and Golanov, Andrey and Belyaev, Mikhail},
	year = {2020},
	pages = {523--532},
}

@inproceedings{sudre_generalised_2017,
	address = {Cham},
	title = {Generalised {Dice} {Overlap} as a {Deep} {Learning} {Loss} {Function} for {Highly} {Unbalanced} {Segmentations}},
	isbn = {978-3-319-67558-9},
	doi = {10.1007/978-3-319-67558-9_28},
	booktitle = {Deep {Learning} in {Medical} {Image} {Analysis} and {Multimodal} {Learning} for {Clinical} {Decision} {Support}},
	volume = {10553},
	series = {Lecture {Notes} in {Computer} {Science}},
	publisher = {Springer},
	author = {Sudre, Carole H. and Li, Wenqi and Vercauteren, Tom and Ourselin, Sebastien and Cardoso, M. Jorge},
	year = {2017},
	pages = {240--248},
}

@misc{rondinella_icpr_2024,
	title = {{ICPR} 2024 {Competition} on {Multiple} {Sclerosis} {Lesion} {Segmentation} -- {Methods} and {Results}},
	url = {https://arxiv.org/abs/2410.07924},
	doi = {10.48550/arXiv.2410.07924},
	publisher = {arXiv},
	author = {Rondinella, Alessia and Guarnera, Francesco and Crispino, Elena and Russo, Giulia and Di Lorenzo, Clara and Maimone, Davide and Pappalardo, Francesco and Battiato, Sebastiano},
	year = {2024},
}

@misc{basaran_seghed_2024,
	title = {{SegHeD}+: {Segmentation} of {Heterogeneous} {Data} for {Multiple} {Sclerosis} {Lesions} with {Anatomical} {Constraints} and {Lesion}-aware {Augmentation}},
	url = {https://arxiv.org/abs/2412.10946},
	doi = {10.48550/arXiv.2412.10946},
	publisher = {arXiv},
	author = {Basaran, Berke Doga and Matthews, Paul M. and Bai, Wenjia},
	year = {2024},
}

@misc{wynen_conflunet_2025,
	title = {{ConfLUNet}: {Multiple} sclerosis lesion instance segmentation in presence of confluent lesions},
	url = {https://arxiv.org/abs/2505.22537},
	doi = {10.48550/arXiv.2505.22537},
	publisher = {arXiv},
	author = {Wynen, Maxence and Gordaliza, Pedro M. and Istasse, Maxime and Stölting, Anna and Maggi, Pietro and Macq, Benoît and Bach Cuadra, Meritxell},
	year = {2025},
}

@misc{molchanova_benchmarking_2025,
	title = {Benchmarking and {Explaining} {Deep} {Learning} {Cortical} {Lesion} {MRI} {Segmentation} in {Multiple} {Sclerosis}},
	url = {https://arxiv.org/abs/2507.12092},
	doi = {10.48550/arXiv.2507.12092},
	publisher = {arXiv},
	author = {Molchanova, Nataliia and Cagol, Alessandro and Ocampo-Pineda, Mario and Lu, Po-Jui and Weigel, Matthias and Chen, Xinjie and Beck, Erin S. and Tsagkas, Charidimos and Reich, Daniel S. and Vanden Bulcke, Colin and Stölting, Anna and Borrelli, Serena and Maggi, Pietro and Depeursinge, Adrien and Granziera, Cristina and Müller, Henning and Gordaliza, Pedro M. and Bach Cuadra, Meritxell},
	year = {2025},
}

@misc{cheng_fmamil_2025,
	title = {{FMaMIL}: {Frequency}-{Driven} {Mamba} {Multi}-{Instance} {Learning} for {Weakly} {Supervised} {Lesion} {Segmentation} in {Medical} {Images}},
	url = {https://arxiv.org/abs/2506.07652},
	doi = {10.48550/arXiv.2506.07652},
	publisher = {arXiv},
	author = {Cheng, Hangbei and Dong, Xiaorong and Liu, Xueyu and Zhang, Jianan and Ma, Xuetao and Wei, Mingqiang and Wang, Liansheng and Chen, Junxin and Wu, Yongfei},
	year = {2025},
}

@misc{basit_rethinking_2026,
	title = {Rethinking {Evaluation} of {Multiple} {Sclerosis} ({MS}) {Lesion} {Segmentation} {Models}},
	url = {https://arxiv.org/abs/2605.09666},
	doi = {10.48550/arXiv.2605.09666},
	publisher = {arXiv},
	author = {Basit, Abdul and Rashid, Ashir and Hanif, Muhammad Abdullah and Shafique, Muhammad},
	year = {2026},
	note = {Accepted at IJCNN 2026},
}

@misc{bouteille_bicc_2026,
	title = {{BiCC}: {Bidirectional} {Connected}-{Component} {Loss} for {Instance}-{Aware} {Segmentation}},
	url = {https://arxiv.org/abs/2609.30223},
	doi = {10.48550/arXiv.2609.30223},
	publisher = {arXiv},
	author = {Bouteille, Luc and Jonske, Frederic and Kleesiek, Jens and Jaus, Alexander},
	year = {2026},
}

@article{filippi_mri_2016,
	title = {{MRI} criteria for the diagnosis of multiple sclerosis: {MAGNIMS} consensus guidelines},
	volume = {15},
	doi = {10.1016/S1474-4422(15)00393-2},
	number = {3},
	journal = {The Lancet Neurology},
	author = {Filippi, Massimo and Rocca, Maria A and Ciccarelli, Olga and De Stefano, Nicola and Evangelou, Nikos and Kappos, Ludwig and Rovira, Alex and Sastre-Garriga, Jaume and Tintorè, Mar and Frederiksen, Jette L and Gasperini, Claudio and Palace, Jacqueline and Reich, Daniel S and Banwell, Brenda and Montalban, Xavier and Barkhof, Frederik},
	year = {2016},
	pages = {292--303},
}

@article{barkhof_2024_2025,
	title = {2024 {MAGNIMS}--{CMSC}--{NAIMS} consensus recommendations on the use of {MRI} for the diagnosis of multiple sclerosis},
	volume = {24},
	doi = {10.1016/S1474-4422(25)00304-7},
	number = {10},
	journal = {The Lancet Neurology},
	author = {Barkhof, Frederik and Reich, Daniel S and Oh, Jiwon and Rocca, Maria A and Li, David K B and Sati, Pascal and others},
	year = {2025},
	pages = {866--879},
}

@article{montalban_diagnosis_2025,
	title = {Diagnosis of multiple sclerosis: 2024 revisions of the {McDonald} criteria},
	volume = {24},
	doi = {10.1016/S1474-4422(25)00270-4},
	number = {10},
	journal = {The Lancet Neurology},
	author = {Montalban, Xavier and Lebrun-Frénay, Christine and Oh, Jiwon and Arrambide, Georgina and Moccia, Marcello and Amato, Maria Pia and others},
	year = {2025},
	pages = {850--865},
}

@incollection{crimi_3d_2019,
	address = {Cham},
	title = {{3D} {MRI} {Brain} {Tumor} {Segmentation} {Using} {Autoencoder} {Regularization}},
	volume = {11384},
	isbn = {978-3-030-11725-2 978-3-030-11726-9},
	url = {https://link.springer.com/10.1007/978-3-030-11726-9_28},
	doi = {10.1007/978-3-030-11726-9_28},
	language = {english},
	urldate = {2026-05-17},
	booktitle = {Brainlesion: {Glioma}, {Multiple} {Sclerosis}, {Stroke} and {Traumatic} {Brain} {Injuries}},
	publisher = {Springer International Publishing},
	author = {Myronenko, Andriy},
	editor = {Crimi, Alessandro and Bakas, Spyridon and Kuijf, Hugo and Keyvan, Farahani and Reyes, Mauricio and Van Walsum, Theo},
	year = {2019},
	series = {Lecture Notes in Computer Science},
	pages = {311--320},
}

@article{wattjes_2021_2021,
  title = {2021 {MAGNIMS}--{CMSC}--{NAIMS} consensus recommendations on the use of {MRI} in patients with multiple sclerosis},
  author = {Wattjes, Mike P. and Ciccarelli, Olga and Reich, Daniel S. and others},
  journal = {The Lancet Neurology},
  volume = {20},
  number = {8},
  pages = {653--670},
  year = {2021},
  doi = {10.1016/S1474-4422(21)00095-8}
}

\clearpage
\appendix
\crefalias{section}{appendix}
\section{Definitions of Evaluation Metrics}
\label{appendix:metrics}

Let $P$ and $G$ denote the sets of predicted and reference lesion voxels of
a test case, and $|\cdot|$ the number of voxels in a set. Let
$\mathcal{G}$ and $\mathcal{P}$ denote the sets of reference and predicted
lesions (6-connected components), and
$\mathcal{G}_{\mathrm{det}} \subseteq \mathcal{G}$ and
$\mathcal{P}_{\mathrm{det}} \subseteq \mathcal{P}$ the reference lesions
that overlap at least one predicted lesion and the predicted lesions that
overlap at least one reference lesion. In all formulas,
$\varepsilon = 10^{-8}$ prevents division by zero. Every metric is computed
per test case and then averaged over cases.

\paragraph{Voxel-level metrics.}
\begin{equation}
\mathrm{DSC} = \frac{2|P \cap G|}{|P| + |G|}.
\end{equation}
With $S_P$ and $S_G$ the surface voxels of the two masks and
$d(p, S_G) = \min_{g \in S_G} \lVert p - g \rVert_2$ in mm,
\begin{equation}
\mathrm{HD}_{95}(P,G) =
\max \left\{
\underset{p \in S_P}{\mathrm{percentile}_{95}}\, d(p, S_G),\;
\underset{g \in S_G}{\mathrm{percentile}_{95}}\, d(g, S_P)
\right\}.
\end{equation}

\paragraph{Lesion-level metrics.}
\begin{equation}
\mathrm{Precision}_{\mathrm{les}} =
\frac{|\mathcal{P}_{\mathrm{det}}|}{|\mathcal{P}| + \varepsilon},
\qquad
\mathrm{Recall}_{\mathrm{les}} =
\frac{|\mathcal{G}_{\mathrm{det}}|}{|\mathcal{G}| + \varepsilon},
\qquad
\mathrm{F1}_{\mathrm{les}} =
\frac{2 \, \mathrm{Precision}_{\mathrm{les}} \, \mathrm{Recall}_{\mathrm{les}}}
{\mathrm{Precision}_{\mathrm{les}} + \mathrm{Recall}_{\mathrm{les}} + \varepsilon}.
\end{equation}
With $\tau = 150$ voxels and
$\mathcal{G}_{\mathrm{small}} = \{ g \in \mathcal{G} : |g| \leq \tau \}$,
\begin{equation}
\mathrm{Recall}_{\mathrm{small}} =
\frac{|\mathcal{G}_{\mathrm{small}} \cap \mathcal{G}_{\mathrm{det}}|}
{|\mathcal{G}_{\mathrm{small}}| + \varepsilon},
\qquad
\text{FN Count} = |\mathcal{G}| - |\mathcal{G}_{\mathrm{det}}|.
\end{equation}
The recall of lesions of at least 3~mm is $\mathrm{Recall}_{\mathrm{les}}$
restricted to reference lesions whose longest diameter is at least 3~mm.
The longest diameter of a lesion is the maximum, over all axial, coronal,
and sagittal slices that intersect it, of the largest distance between two
of its voxel centers in that slice, plus one voxel width (the mean in-plane
spacing), so that it is measured from edge to edge.

\paragraph{Error-volume metrics.}
Let $\mathrm{FP} = P \setminus G$ denote the false-positive voxels and
$s_x, s_y, s_z$ the voxel spacing. The false-positive volume is
\begin{equation}
V_{\mathrm{FP}} = |\mathrm{FP}| \cdot s_x s_y s_z.
\end{equation}
The voxels of $\mathrm{FP}$ are grouped into connected components; a
component is a boundary extension if its minimum distance to the reference
mask is at most 2~mm, and a blob otherwise. With $\mathcal{B}$ the set of
blobs,
\begin{equation}
\mathrm{FP}_{\mathrm{blob}}^{\mathrm{frac}}
= \frac{\sum_{c \in \mathcal{B}} |c|}{|\mathrm{FP}| + \varepsilon},
\qquad
V_{\mathrm{FN}}^{\mathrm{frac}}
= \frac{\sum_{g \in \mathcal{G} \setminus \mathcal{G}_{\mathrm{det}}} |g|}
{|G| + \varepsilon}.
\end{equation}

\section{Sensitivity to Loss Weights}
\label{appendix:catmil-loss-weights}

This section analyzes the impact of the weighting coefficients
$\lambda_{\text{CAT}}$ and $\lambda_{\text{MIL}}$. The configurations are
named CATMIL$xy$, where $\lambda_{\text{CAT}}=0.x$ and
$\lambda_{\text{MIL}}=0.y$; CATMIL0102 is the configuration used in the
main experiments. Its weights were fixed before this analysis was run, and
the analysis was not used to select them: it is a post hoc check of how
sensitive the results are to the choice of weights. Unlike in the other
experiments, each configuration in this analysis was trained once on the
complete development set, without cross-validation, and evaluated on the
same test set. The values of CATMIL0102 therefore differ from those of
CATMIL in \cref{tab:loss_comparison_small_lesion,tab:ablation_cat_mil},
which are means over the five cross-validation models.

\begin{table}[H]
\centering
\caption{Sensitivity of the CATMIL objective to $\lambda_{\text{CAT}}$ and
$\lambda_{\text{MIL}}$ on the MSLesSeg test set.}
\label{tab:ablation_lambda}
\begin{tabular}{lccccc}
\toprule
Model & DSC & HD95 $\downarrow$ & Lesion F1 & Small Recall & FP Vol $\downarrow$ \\
\midrule
CATMIL0101 & 0.7675 & 9.3881 & \underline{0.7720} & 0.8514 & 2292.17 \\
CATMIL0102 & \textbf{0.7954} & \textbf{5.9771} & 0.7452 & \underline{0.8711} & \underline{1345.83} \\
CATMIL0201 & \underline{0.7921} & \underline{6.7964} & \textbf{0.7754} & \textbf{0.8784} & \textbf{1233.92} \\
CATMIL0202 & 0.7663 & 9.4507 & 0.7632 & 0.8699 & 1729.42 \\
\bottomrule
\end{tabular}
\end{table}

The configuration used in the main experiments (CATMIL0102), in which the
MIL term has the larger weight, gives the best global segmentation
performance, with the highest DSC (0.7954) and lowest HD95 (5.9771~mm), but
also the lowest lesion-wise F1 (0.7452). Exchanging the two weights
(CATMIL0201) improves lesion-level behavior, with the highest lesion F1
(0.7754), the highest small lesion recall (0.8784), and the lowest FP
volume (1233.92~mm$^3$), at the cost of slightly lower DSC and boundary
accuracy. The two configurations with equal weights (CATMIL0101 and
CATMIL0202) are clearly worse in DSC, HD95, and FP volume. Small lesion
recall stays between 0.85 and 0.88 in all four configurations, so the
detection gain does not depend on a precise choice of the weights in this
range; the ratio of $\lambda_{\text{CAT}}$ to $\lambda_{\text{MIL}}$ mainly
shifts the balance between voxel-level accuracy and lesion-wise precision.

\section{Lesion-Profile Characteristics of Test Cases}
\label{appendix:test-case-lesion-profile}

Each MSLesSeg test case is summarized using the ground-truth lesion mask
(\cref{tab:test_case_lesion_profile}). Lesions are counted by applying
connected-component analysis with 6-connectivity to the binary lesion
mask, the same definition of a lesion as in the proposed loss and in the
evaluation metrics; all components are counted regardless of their size.
Lesion volume is the number of voxels inside each lesion multiplied by the
voxel volume; at the 1~mm isotropic spacing of MSLesSeg, one voxel
corresponds to 1~mm$^3$. The total volume is the sum of all lesion volumes
in one case, and the median volume represents the typical lesion size in
that case.

Lesion contrast is computed locally on the FLAIR image after z-score
normalization within the brain. For each lesion, a surrounding background
ring is created within 3~mm around the lesion, excluding the lesion itself.
The contrast value is the absolute difference between the median intensity
inside the lesion and the median intensity in this ring, divided by the
median absolute deviation of the ring intensity. A higher value indicates
that the lesion is more distinguishable from its local background.

Boundary sharpness is computed from the image gradient around the lesion
boundary. The normalized FLAIR image is smoothed with a Gaussian filter of
1~mm, and the sharpness is the median gradient magnitude in a band of one
voxel on each side of the lesion boundary. A higher value indicates a
clearer lesion boundary.

Contrast and sharpness are not estimated reliably for the smallest lesions
and are therefore computed only for lesions of at least 10 voxels. The
contrast and sharpness in \cref{tab:test_case_lesion_profile} are means
over these lesions, whereas the lesion count and the volume statistics
include all lesions.

\begin{table}[H]
\centering
\caption{Lesion-profile characteristics of the MSLesSeg test cases. N:
number of lesions; $\leq$50 and $\leq$100: percentage of lesions of at most
50 and 100~mm$^3$.}
\label{tab:test_case_lesion_profile}
\begin{tabular}{lrrrrrrr}
\toprule
\textbf{Case} & \textbf{N} & \textbf{Total vol.} & \textbf{Median vol.}
& \textbf{$\leq$50} & \textbf{$\leq$100} & \textbf{Contrast} & \textbf{Sharpness} \\
& & \textbf{(mm$^3$)} & \textbf{(mm$^3$)} & \textbf{(\%)} & \textbf{(\%)} & & \\
\midrule
P13\_T1 & 39 & 9677 & 83.0 & 10.3 & 53.8 & 4.11 & 8.41 \\
P13\_T2 & 61 & 10268 & 68.0 & 45.9 & 65.6 & 3.85 & 8.34 \\
P26\_T1 & 43 & 6639 & 106.0 & 20.9 & 46.5 & 3.77 & 9.16 \\
P2\_T1  & 16 & 2261 & 92.5 & 31.2 & 50.0 & 4.56 & 9.31 \\
P2\_T2  & 20 & 3286 & 87.5 & 20.0 & 55.0 & 3.76 & 10.30 \\
P2\_T3  & 14 & 2623 & 134.0 & 14.3 & 28.6 & 4.61 & 11.68 \\
P2\_T4  & 18 & 2940 & 87.0 & 16.7 & 55.6 & 4.45 & 10.86 \\
P47\_T1 & 59 & 18736 & 67.0 & 42.4 & 71.2 & 4.26 & 9.92 \\
P49\_T1 & 30 & 34709 & 83.5 & 30.0 & 56.7 & 3.65 & 8.55 \\
P49\_T2 & 19 & 7763 & 173.0 & 0.0 & 26.3 & 3.34 & 7.38 \\
P52\_T1 & 28 & 3306 & 70.5 & 42.9 & 67.9 & 5.24 & 11.02 \\
P52\_T2 & 14 & 2843 & 147.0 & 7.1 & 42.9 & 4.04 & 6.91 \\
\bottomrule
\end{tabular}
\end{table}

\section{Additional Lesion-Level Analyses}
\label{appendix:lesion-level-analyses}

This section reports lesion-level analyses that complement the size
analysis in the main text. All analyses use the same lesion definition as
the evaluation metrics: 6-connected components of the reference mask, with
a lesion counted as detected if any predicted voxel overlaps it. Values are
means over the five fold models.

\paragraph{Results after removing small components.}
\Cref{tab:loss_comparison_small_lesion_pp5} repeats the MSLesSeg comparison of
\cref{tab:loss_comparison_small_lesion} after removing predicted components
smaller than 5 voxels (PP5). Paired differences are given in
\cref{tab:paired_differences}.

\begin{table}[htbp]
\centering
\caption{Comparison of CATMIL with standard losses and the instance-level
loss CC-DiceCE on the MSLesSeg test set after post-processing with a
minimum connected-component size of 5 voxels (PP5). Best values are shown
in \textbf{bold}; second-best values are \underline{underlined}.}
\label{tab:loss_comparison_small_lesion_pp5}
\resizebox{\textwidth}{!}{%
\begin{tabular}{lccccc}
\toprule
\textbf{Metric}
& \textbf{CATMIL (ours)}
& \textbf{DiceCE}
& \textbf{Tversky}
& \textbf{FocalTversky}
& \textbf{CC-DiceCE} \\
\midrule
Small Lesion Recall $\uparrow$
& \textbf{0.8554 $\pm$ 0.01}
& 0.7853 $\pm$ 0.03
& \underline{0.8223 $\pm$ 0.02}
& 0.8196 $\pm$ 0.02
& 0.8148 $\pm$ 0.01 \\
FN Count $\downarrow$
& \textbf{3.1333 $\pm$ 0.21}
& 5.2333 $\pm$ 0.52
& \underline{4.0667 $\pm$ 0.24}
& 4.0833 $\pm$ 0.24
& 4.1833 $\pm$ 0.30 \\
FN Volume Fraction $\downarrow$
& \textbf{0.0238 $\pm$ 0.004}
& 0.0353 $\pm$ 0.01
& 0.0277 $\pm$ 0.01
& \underline{0.0270 $\pm$ 0.01}
& 0.0369 $\pm$ 0.01 \\
\midrule
DSC $\uparrow$
& \textbf{0.7842 $\pm$ 0.01}
& 0.7797 $\pm$ 0.01
& 0.7707 $\pm$ 0.01
& \underline{0.7803 $\pm$ 0.01}
& 0.7781 $\pm$ 0.01 \\
HD95 (mm) $\downarrow$
& \textbf{8.1028 $\pm$ 1.76}
& 9.0895 $\pm$ 1.37
& 10.3585 $\pm$ 0.21
& \underline{8.2297 $\pm$ 1.95}
& 9.8234 $\pm$ 1.52 \\
Lesion F1 $\uparrow$
& 0.8415 $\pm$ 0.01
& \textbf{0.8586 $\pm$ 0.01}
& 0.8532 $\pm$ 0.01
& \underline{0.8582 $\pm$ 0.005}
& 0.8387 $\pm$ 0.01 \\
FP Volume (mm$^3$) $\downarrow$
& \underline{1515 $\pm$ 244}
& 1815 $\pm$ 654
& 2618 $\pm$ 558
& 2279 $\pm$ 710
& \textbf{1505 $\pm$ 290} \\
\bottomrule
\end{tabular}%
}
\end{table}

\begin{figure}[htbp]
    \centering
    \includegraphics[width=\textwidth]{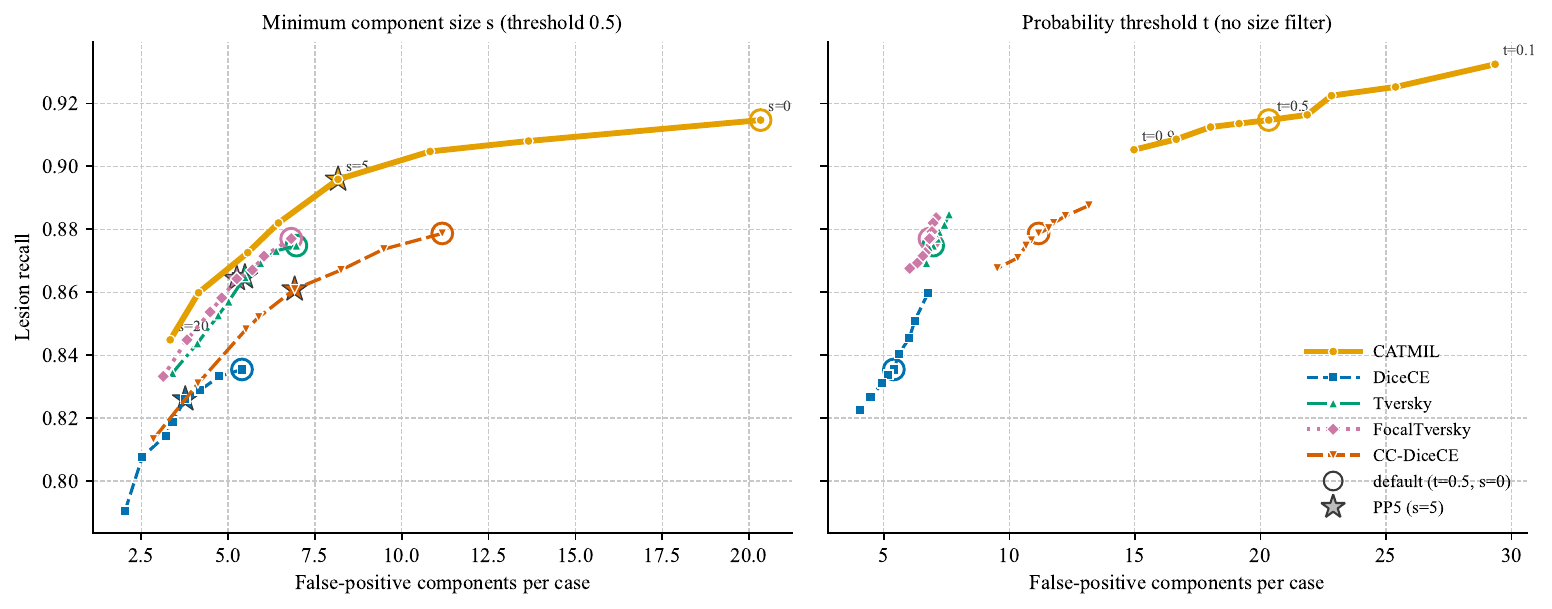}
    \caption{Lesion recall against false-positive components per case on
    the MSLesSeg test set. Left: predicted components smaller than $s$
    voxels are removed ($s = 0, 2, 3, 5, 8, 10, 15, 20$) at probability
    threshold 0.5. Right: the probability threshold $t$ is varied from 0.1
    to 0.9 without filtering. Open circles: unfiltered predictions at
    $t = 0.5$; stars: PP5 ($s = 5$). A false-positive component is a
    predicted connected component that does not overlap any reference
    lesion. Recall is pooled over the test lesions; values are means over
    the five fold models.}
    \label{fig:operating_curve}
\end{figure}

\paragraph{Operating points under component filtering.}
PP5 uses a single filter size. \Cref{fig:operating_curve} shows how lesion
recall and the number of false-positive components per case change when
the filter size is varied. In the left panel, predicted components smaller
than $s$ voxels are removed, with $s$ from 0 to 20; in the right panel, the
probability threshold $t$ is varied from 0.1 to 0.9 without filtering. A
curve that lies higher and further to the left detects more lesions for
the same number of false alarms.

The CATMIL curve lies above the baseline curves across the whole range of
filter sizes. Without filtering, CATMIL produces more false-positive
components than the baselines (20.3 per case, against 5.4--7.0 for the
voxel-level losses and 11.2 for CC-DiceCE), which is the cause of its lower
Lesion F1. These components are small and are
removed at little cost in recall. With $s = 20$, CATMIL detects 84.5\% of
lesions at 3.3 false-positive components per case, more lesions with fewer
false alarms than unfiltered DiceCE (83.5\% at 5.4). For any false-alarm
budget of at least 3.3 components per case, a filtered CATMIL model
therefore detects more lesions than any of the baselines. Raising the
probability threshold is less effective than a size filter, because most
of the false-positive components are confident predictions rather than
borderline ones. The margin over the Tversky-based losses at a fixed
false-alarm level is small, however, and is not supported by the
patient-level analysis at matched precision (\cref{tab:matched_precision}).

\begin{figure}[htbp]
    \centering
    \includegraphics[width=\textwidth]{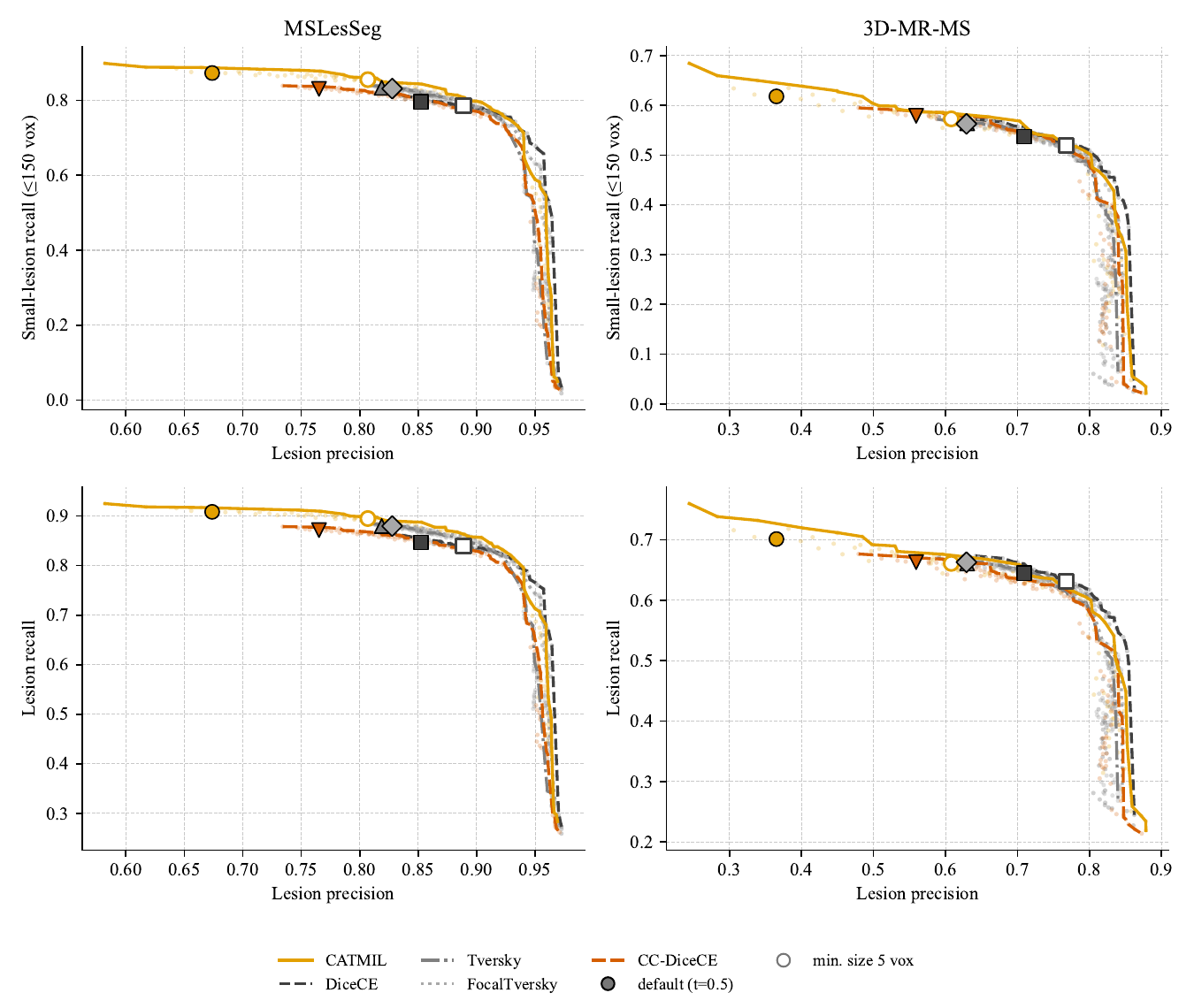}
    \caption{Lesion recall against lesion-wise precision for all
    combinations of probability threshold ($t = 0.1, \dots, 0.9$) and
    minimum component size ($s = 1$ to 170 voxels). Top: small-lesion
    recall ($\leq 150$ voxels); bottom: recall of all lesions. Faint
    points: individual settings; lines: settings not outperformed in both
    precision and recall by another setting of the same loss. Filled
    markers: default output ($t = 0.5$, $s = 1$); open markers: PP5
    ($t = 0.5$, $s = 5$), shown for CATMIL and DiceCE. Metrics are computed
    per case and averaged over cases and the five fold models.}
    \label{fig:operating_grid}
\end{figure}

\paragraph{Joint variation of threshold and filter size.}
\Cref{fig:operating_grid} varies the probability threshold and the minimum
component size jointly and plots lesion recall against lesion-wise
precision, the two quantities that make up lesion-wise F1. Each faint point
is one setting; the lines connect the settings that are not outperformed
in both precision and recall by another setting of the same loss. On
MSLesSeg, the CATMIL frontier lies above those of DiceCE and CC-DiceCE over
most of the precision range and close to those of the Tversky-based losses,
which is the pattern quantified in \cref{tab:matched_precision}. On 3D-MR-MS, the
default output of CATMIL (filled circle) lies far to the left of its
frontier: its precision of 0.37 is due to the many small components, and
the frontier reaches the precision of the baselines with similar recall.
In this analysis, the minimum component size is given in voxels on both
datasets. Expressing it in mm$^3$ instead, that is, removing components of
the same physical volume on both datasets, transfers poorly: because the
voxels of 3D-MR-MS are 5.7 times smaller, a filter of 15~mm$^3$ removes
85 voxels and with them most of the small lesions. The false-positive
components of CATMIL are therefore small in voxels rather than in
millimeters, and the filter size should be chosen for each image
resolution.

\begin{figure}[htbp]
    \centering
    \includegraphics[width=0.85\textwidth]{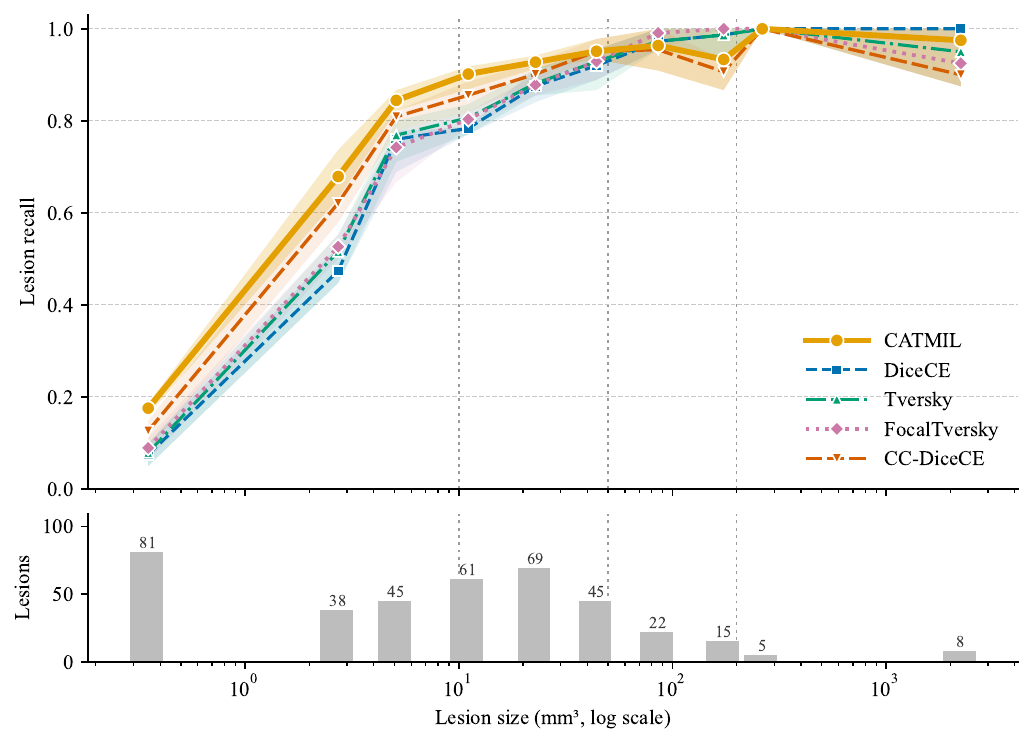}
    \caption{Lesion recall against lesion size on the 3D-MR-MS test set.
    Layout as in \cref{fig:recall_vs_size_curve}.}
    \label{fig:recall_vs_size_curve_3dmrms}
\end{figure}

\begin{figure}[htbp]
    \centering
    \includegraphics[width=\textwidth]{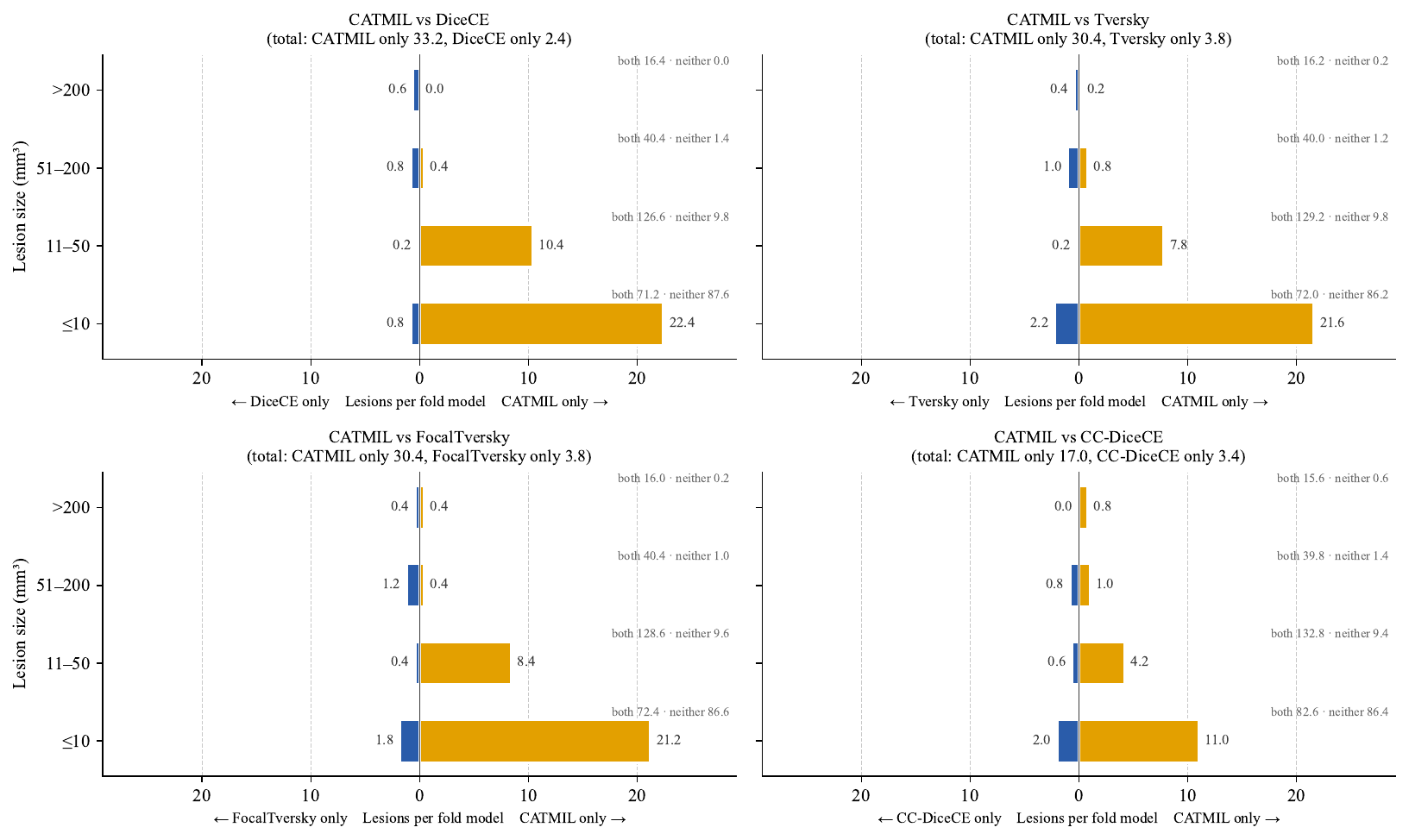}
    \caption{Lesions detected by only one of two losses on the 3D-MR-MS
    test set, by lesion size. Layout as in \cref{fig:lesion_fate}.}
    \label{fig:lesion_fate_3dmrms}
\end{figure}

\begin{figure}[htbp]
    \centering
    \includegraphics[width=\textwidth]{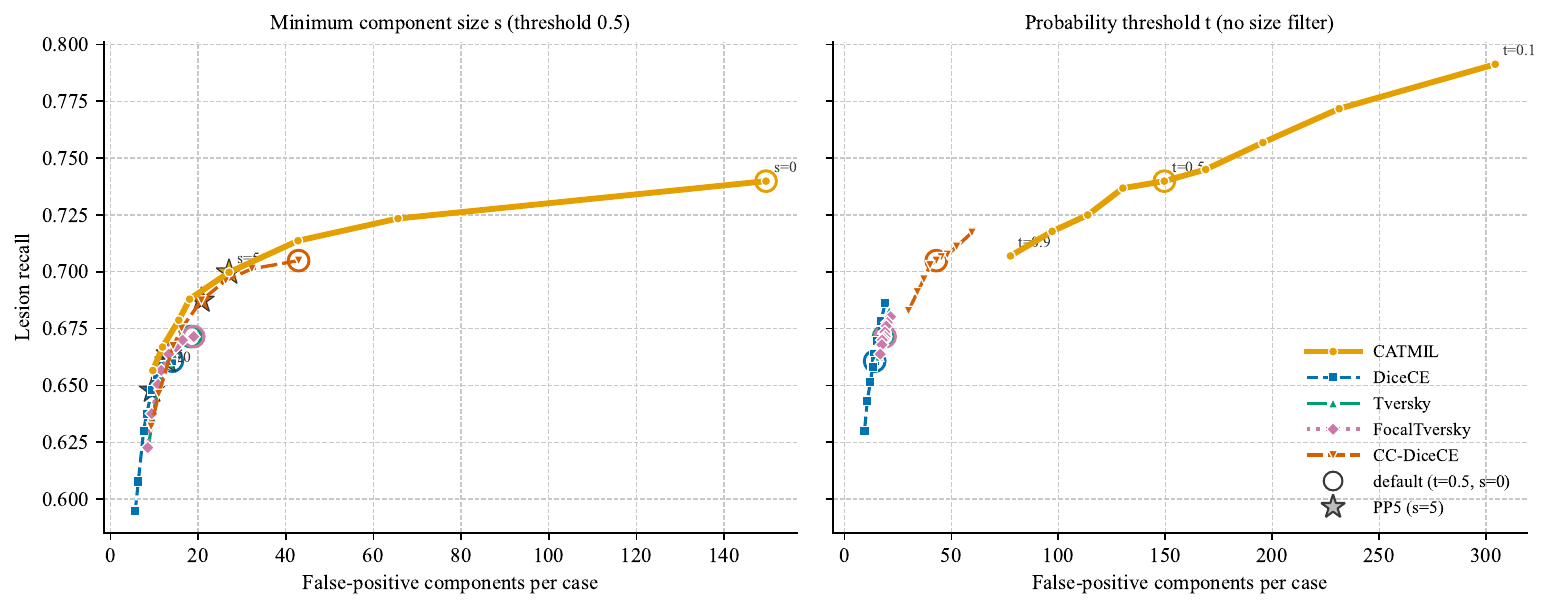}
    \caption{Lesion recall against false-positive components per case on
    the 3D-MR-MS test set. Layout as in \cref{fig:operating_curve}.}
    \label{fig:operating_curve_3dmrms}
\end{figure}

\begin{figure}[htbp]
    \centering
    \includegraphics[width=0.75\textwidth]{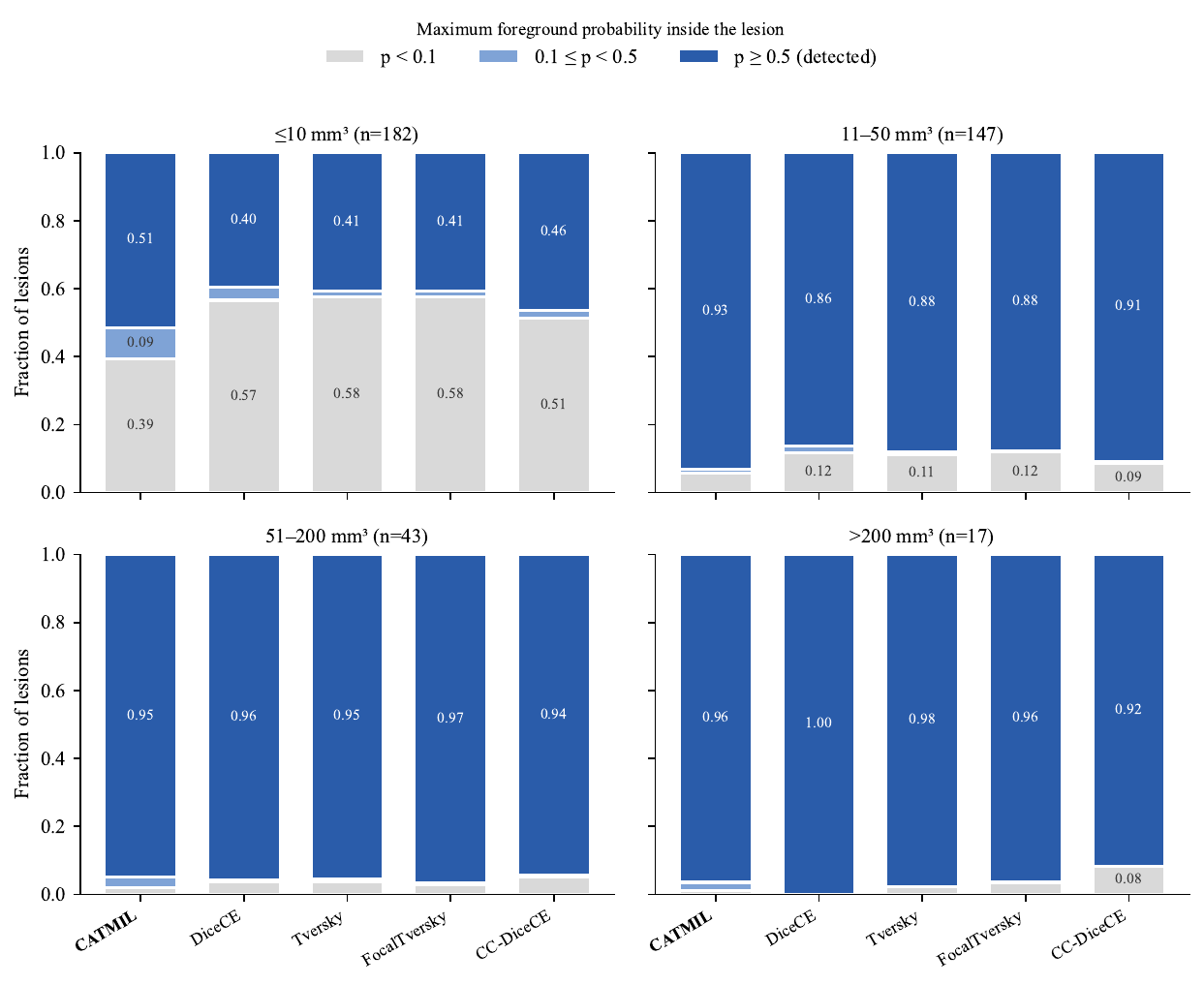}
    \caption{Maximum foreground probability inside each reference lesion on
    the 3D-MR-MS test set, by lesion size. Layout as in
    \cref{fig:max_prob_by_size}.}
    \label{fig:max_prob_by_size_3dmrms}
\end{figure}

\paragraph{Lesion-level analyses on the 3D-MR-MS dataset.}
\Cref{fig:recall_vs_size_curve_3dmrms,fig:lesion_fate_3dmrms,fig:operating_curve_3dmrms,fig:max_prob_by_size_3dmrms}
repeat the analyses of
\cref{fig:recall_vs_size_curve,fig:lesion_fate,fig:operating_curve,fig:max_prob_by_size}
on the 3D-MR-MS dataset, and are read in the same way. Because the voxels
of this dataset are anisotropic ($0.47 \times 0.47 \times 0.8$~mm), lesion
sizes are given in mm$^3$ rather than in voxels.

The recall curve (\cref{fig:recall_vs_size_curve_3dmrms}) reproduces the
main result. CATMIL detects 51\% of lesions of at most 10~mm$^3$, compared
with 40--41\% for the voxel-level losses and 47\% for CC-DiceCE, and 93\%
of lesions of 11--50~mm$^3$, compared with 86--88\% and 91\%. Small lesions
are far more frequent in this dataset: 182 of the 389 reference lesions are
at most 10~mm$^3$, so the gain in this range accounts for most of the
overall increase in lesion recall, from 0.66--0.67 for the voxel-level
losses and 0.71 for CC-DiceCE to 0.74. Above 100~mm$^3$, CATMIL detects
slightly fewer lesions than the voxel-level losses (95\% against 98--99\%),
a difference of about one of the 34 lesions in this range, and slightly
more than CC-DiceCE (93\%).

The paired comparison (\cref{fig:lesion_fate_3dmrms}) is again one-sided.
CATMIL detects 33.2 lesions per fold model that DiceCE misses, against 2.4
in the opposite direction, 30.4 against 3.8 for each of the Tversky-based
losses, and 17.0 against 3.4 for CC-DiceCE. About 97\% of the lesions
recovered by CATMIL relative to the voxel-level losses, and 89\% of those
recovered relative to CC-DiceCE, are at most 50~mm$^3$.

The operating curve (\cref{fig:operating_curve_3dmrms}) shows that the
component-level cost of CATMIL is larger on this dataset. Without
filtering, CATMIL produces 150 false-positive components per case, compared
with 14--19 for the voxel-level losses and 43 for CC-DiceCE, which explains its low Lesion F1 in
\cref{tab:loss_comparison_small_lesion}. The size filter again removes most of
them: PP5 reduces the count to 27 per case with a recall of 0.70, and a
filter of 20 voxels reduces it to 9.7 per case with a recall of 0.657. At
this setting, CATMIL has about as many false-positive components as DiceCE
after PP5 (9.5 per case) and a higher recall (0.648 for DiceCE). The filter
size required to control false positives is therefore dataset dependent,
and a larger filter is needed at the finer in-plane resolution of
3D-MR-MS.

The probability levels (\cref{fig:max_prob_by_size_3dmrms}) show the same
mechanism as on MSLesSeg. For lesions of at most 10~mm$^3$, the fraction
with no response falls from 57--58\% for the voxel-level losses and 51\%
for CC-DiceCE to 39\% for CATMIL.

\section{Robustness to the Detection Criterion}
\label{appendix:strict-detection}

All lesion-level results in the main text count a reference lesion as
detected if any predicted voxel overlaps it. Under this criterion, a
prediction that covers a single voxel of a lesion counts as a detection,
and the case-based analysis shows that CATMIL often covers only part of
the lesions it recovers. \Cref{tab:strict_detection} therefore recomputes
small-lesion recall with stricter criteria, which require a minimum
fraction of the lesion voxels to be covered by the default segmentation
($p > 0.5$).

\begin{table}[htbp]
\centering
\caption{Small-lesion recall ($\leq 150$ voxels) under detection criteria
of increasing strictness. A lesion counts as detected if the prediction
covers any of its voxels, or at least the stated fraction of them. Recall
is pooled over the reference lesions of the test set and averaged over the
five fold models, so the any-overlap values differ slightly from the
case-averaged Small Lesion Recall in the main tables. $\Delta$: CATMIL
minus DiceCE or CC-DiceCE with 95\% patient-level bootstrap CI; differences
whose CI excludes zero are shown in \textbf{bold}.}
\label{tab:strict_detection}
\resizebox{\textwidth}{!}{%
\begin{tabular}{llcccccll}
\toprule
\textbf{Dataset} & \textbf{Criterion} & \textbf{CATMIL} & \textbf{DiceCE} & \textbf{Tversky} & \textbf{FocalTversky} & \textbf{CC-DiceCE} & $\Delta$ vs.\ DiceCE & $\Delta$ vs.\ CC-DiceCE \\
\midrule
MSLesSeg & any overlap & 0.883 & 0.779 & 0.829 & 0.833 & 0.841 & \textbf{+0.104 [+0.046, +0.176]} & \textbf{+0.042 [+0.024, +0.062]} \\
 & $\geq$ 10\% & 0.865 & 0.767 & 0.818 & 0.818 & 0.826 & \textbf{+0.098 [+0.039, +0.170]} & \textbf{+0.039 [+0.020, +0.060]} \\
 & $\geq$ 25\% & 0.836 & 0.743 & 0.802 & 0.799 & 0.799 & \textbf{+0.093 [+0.032, +0.162]} & \textbf{+0.037 [+0.012, +0.062]} \\
 & $\geq$ 50\% & 0.760 & 0.679 & 0.750 & 0.748 & 0.725 & \textbf{+0.081 [+0.029, +0.138]} & \textbf{+0.035 [+0.007, +0.066]} \\
\midrule
3D-MR-MS & any overlap & 0.654 & 0.548 & 0.564 & 0.562 & 0.614 & \textbf{+0.107 [+0.046, +0.129]} & \textbf{+0.041 [+0.011, +0.080]} \\
 & $\geq$ 10\% & 0.630 & 0.538 & 0.555 & 0.556 & 0.602 & \textbf{+0.092 [+0.029, +0.114]} & +0.028 [$-$0.004, +0.062] \\
 & $\geq$ 25\% & 0.595 & 0.509 & 0.534 & 0.538 & 0.584 & \textbf{+0.086 [+0.009, +0.108]} & +0.011 [$-$0.034, +0.046] \\
 & $\geq$ 50\% & 0.550 & 0.449 & 0.493 & 0.500 & 0.535 & \textbf{+0.101 [+0.028, +0.129]} & \textbf{+0.015 [+0.002, +0.046]} \\
\bottomrule
\end{tabular}%
}
\end{table}

The gain over DiceCE does not depend on the lenient criterion. On both
datasets it stays between 8 and 11 percentage points at every coverage
level, and its CI excludes zero in every case. The margin over the
Tversky-based losses narrows as the criterion becomes stricter: at 50\%
coverage on MSLesSeg, CATMIL detects 76.0\% of small lesions against
74.8--75.0\%. The gain over CC-DiceCE stays between 3.5 and 4.2 points at
every coverage level on MSLesSeg, with CIs that exclude zero, but on
3D-MR-MS it shrinks from 4.1 points under any overlap to 1.1--2.8 points
under the coverage criteria, with CIs that include zero at 10\% and 25\%.
This agrees with the case-based analysis, in which CATMIL
recovers lesions that the baselines miss but often covers only part of
them; the lesion-level term ensures that a lesion is detected, whereas its
full extent still depends on the voxel-wise terms. Lesion-wise precision
cannot be recomputed with a coverage criterion from the stored lesion
tables, so this analysis is limited to recall.

\section{Gradient and Boundedness Properties of the CATMIL Objective}
\label{appendix:catmil-properties}

This section derives the properties of the CAT and MIL terms stated in
\cref{sec:method}, using the notation of that section. The components $C_k$
and the weights $w_i$ are computed from the ground-truth mask and are
constants with respect to the predictions $p_i$.

\paragraph{Gradient of the CAT term.}
Let
\begin{equation}
A = \mathrm{TP}_{w}+\delta,
\qquad
B = A + \alpha\, \mathrm{FP}_{w} + \beta\, \mathrm{FN}_{w},
\end{equation}
so that $\mathcal{L}_{\mathrm{CAT}} = 1 - A/B$. For a foreground voxel
$j \in C_k$, $\partial \mathrm{TP}_{w}/\partial p_j = w_j$,
$\partial \mathrm{FP}_{w}/\partial p_j = 0$, and
$\partial \mathrm{FN}_{w}/\partial p_j = -w_j$, hence
$\partial A/\partial p_j = w_j$ and
$\partial B/\partial p_j = (1-\beta)\, w_j$. The quotient rule gives
\begin{equation}
\frac{\partial \mathcal{L}_{\mathrm{CAT}}}{\partial p_j}
=
-\frac{w_j B - (1-\beta)\, w_j A}{B^2}
=
-c\, w_j ,
\qquad
c
=
\frac{\beta A + \alpha\, \mathrm{FP}_{w} + \beta\, \mathrm{FN}_{w}}{B^2}
> 0 .
\end{equation}
The factor $c$ is the same for all foreground voxels, so the gradient at a
lesion voxel is scaled only by its weight
$w_j = (|C_k|+\varepsilon_{c})^{-\gamma}$. Summing over the component, the
total gradient magnitude received by lesion $C_k$ is
\begin{equation}
\sum_{j \in C_k}
\left| \frac{\partial \mathcal{L}_{\mathrm{CAT}}}{\partial p_j} \right|
=
c\, |C_k|\, (|C_k|+\varepsilon_{c})^{-\gamma},
\end{equation}
which is proportional to $|C_k|$ for $\gamma=0$ and equal to
$c\,\rho_k < c$ for $\gamma=1$, with $\rho_k$ the lesion weight of
\cref{eq:sl_cat_lesion_weight}.

For a background voxel $j$ with $g_j=0$, only $\mathrm{FP}_{w}$ depends on
$p_j$, so $\partial A/\partial p_j = 0$ and
$\partial B/\partial p_j = \alpha\, w_{\mathrm{bg}}$, and
\begin{equation}
\frac{\partial \mathcal{L}_{\mathrm{CAT}}}{\partial p_j}
=
\frac{\alpha\, w_{\mathrm{bg}}\, A}{B^2}
\geq 0 .
\end{equation}
Increasing the predicted probability at a background voxel therefore
increases the loss.

\paragraph{Gradient of the MIL term.}
With the clipping of \cref{eq:sl_mil_loss}, the contribution of lesion
$C_k$ is $-\log\bigl(\max(s_k,\varepsilon)\bigr)$ and remains finite. Let
$m_k = \arg\max_{i \in C_k} p_i$ be the maximal voxel of lesion $C_k$.
Away from ties,
\begin{equation}
\frac{\partial \mathcal{L}_{\mathrm{MIL}}}{\partial p_j}
=
\begin{cases}
-\dfrac{1}{K s_k},
& \text{if } j=m_k \text{ for some } k \text{ and } s_k>\varepsilon, \\[2mm]
0, & \text{otherwise.}
\end{cases}
\end{equation}
The gradient is concentrated on one voxel per lesion, its magnitude does
not depend on $|C_k|$, and it grows as the strongest response $s_k$
decreases. When several voxels share the maximum, the hard maximum is not
differentiable but remains subdifferentiable, and automatic differentiation
assigns the gradient to a maximal voxel, as in max-pooling.

\paragraph{Boundedness and continuity.}
Since $\mathrm{TP}_{w}$, $\mathrm{FP}_{w}$, $\mathrm{FN}_{w}$, $\alpha$, and
$\beta$ are non-negative and $\delta>0$, $0 < A \leq B$, and therefore
$0 \leq \mathcal{L}_{\mathrm{CAT}} < 1$. Since
$\varepsilon \leq \max(s_k,\varepsilon) \leq 1$,
$0 \leq \mathcal{L}_{\mathrm{MIL}} \leq -\log \varepsilon$. For
non-negative coefficients the auxiliary part of the objective thus
satisfies
\begin{equation}
0
\leq
\lambda_{\mathrm{CAT}}\, \mathcal{L}_{\mathrm{CAT}}
+
\lambda_{\mathrm{MIL}}\, \mathcal{L}_{\mathrm{MIL}}
\leq
\lambda_{\mathrm{CAT}} - \lambda_{\mathrm{MIL}} \log \varepsilon .
\end{equation}
Both terms are continuous in the predictions. The CAT term is a ratio of
linear functions with a strictly positive denominator and is differentiable
everywhere. The MIL term is differentiable except at ties of the maximum
and at the clipping point $s_k=\varepsilon$, which is the same piecewise
behavior as that of max-pooling and does not prevent gradient-based
training.

\end{document}